\documentclass[a4paper]{article}

\usepackage[english]{babel}
\usepackage[utf8]{inputenc}
\usepackage{amsmath}
\usepackage{graphicx}
\usepackage[colorinlistoftodos]{todonotes}
\usepackage{amssymb,amsthm}
\usepackage{hyperref}
\usepackage{verbatim}
\usepackage{multirow}
\usepackage{caption}

\DeclareMathOperator*{\argmax}{arg\,max}
\DeclareMathOperator*{\argmin}{arg\,min}

\title{A Primer on Domain Adaptation\\
\begin{large}
Theory and Applications
\end{large}
}

\author{
Pirmin Lemberger, Ivan Panico\\
{\small \texttt{p.lemberger@groupeonepoint.com,      i.panico@groupeonepoint.com}}\\
\\
   {\small \textit{onepoint}}\\
   {\small 29 rue des Sablons, 75116 Paris} \\
   {\small \texttt{groupeonepoint.com}}\\ \\
   }
\date{\today}

\begin{document}
\maketitle

\newcommand{\bx}{\mathbf{x}}
\newcommand{\by}{\mathbf{y}}
\newcommand{\bz}{\mathbf{z}}
\newcommand{\xfeatures}{x_1 \dots, x_p}
\newcommand{\Xset}{\mathcal{X}}
\newcommand{\Yset}{\mathcal{Y}}
\newcommand{\R}{\mathbb{R}}
\newcommand{\E}{\mathbb{E}}


\begin{abstract}
Standard supervised machine learning assumes that the distribution of the source samples used to train an algorithm is the same as the one of the target samples on which it is supposed to make predictions. However, as any data scientist will confirm, this is hardly ever the case in practice. The set of statistical and numerical methods that deal with such situations is known as domain adaptation, a field with a long and rich history. The myriad of methods available and the unfortunate lack of a clear and universally accepted terminology can however make the topic rather daunting for the newcomer. Therefore, rather than aiming at completeness, which leads to exhibiting a tedious catalog of methods, this pedagogical review aims at a coherent presentation of four important special cases: (1) \emph{prior shift}, a situation in which training samples were selected according to their labels without any knowledge of their actual distribution in the target, (2) \emph{covariate shift} which deals with a situation where training examples were picked according to their features but with some selection bias, (3) \emph{concept shift} where the dependence of the labels on the features defers between the source and the target, and last but not least (4) \emph{subspace mapping} which deals with a situation where features in the target have been subjected to an unknown distortion with respect to the source features. In each case we first build an intuition, next we provide the appropriate mathematical framework and eventually we describe a practical application.
\\ 
\end{abstract}

\section{Let's clear up the fog}
Machine learning (ML) aims at making statistical predictions for some phenomena given appropriate data in sufficient quantity. In supervised machine learning this data come as a set of observations where some target quantity, or category, or label, is dependent on some measurable features. In the standard setting it is assumed that all observations are independent and sampled from one and the same distribution. The role of an ML algorithm is then to convert this training data into a function that accurately predicts the label of an unseen observations given their features. Classical theory of ML assumes that the new observations from the test set, for which we want to make predictions are drawn from the same population as those from the training set. This however is an ideal situation rarely met in practice. The training set is then said to be biased with respect to the test set. Depending on the situation, this bias can either be known or unknown as we shall see. Domain adaptation (DA), which is the subject of this review, is a collection of methods that aims at compensating somehow for the statistical asymmetry between the train set and the test set. As a matter of fact, domain adaptation is everything but a new topic in ML. It has indeed been the focus of numerous research papers and reviews in the past (see \cite{adust_outputs,intro_to_DA} and references therein). There are several reasons however that make it difficult for a newcomer to build a coherent overview of domain adaptation:
\begin{itemize}
\item Research papers on DA which treat the subject within some specific discipline use a specialized vocabulary and concepts that will not apply elsewhere.
\item Even for research on DA that is no specialized to some specific discipline there is no universal terminology agreement for referring to different types of domain adaptation. This often creates confusion.
\item DA is sometimes confused with transfer learning (TL) which is currently a hot topic in NLP and Deep Learning in general. While TL deals with transferring knowledge gained on one task to use it on another\footnote{Let us recall that transfer learning consist in reusing  knowledge gained by training a model on a generic task, typically using large amounts of data, to build another models that will require far less training data to make predictions on another more specialized tasks (which can be quite different from the generic one). One strategy is to fine-tune the generic model by training it on using the limited amount of available specialized data. Another is to extract features from the lower layers of a neural network in a generic model to reuse them in the context of the specialized task. Modern deep learning models like BERT \cite{BERT_paper} use this strategy to build models for an array of NLP tasks by initially training them on large corpora of text to learn simple language models.}, DA, which is our focus, deals with one single task for which the training and test observations have different statistical properties.
\item The literature which deals with DA from a rigorous statistical learning perspective is sometimes heavy on mathematical technicalities which makes it a difficult read for the more hands-on data scientists.
\item On the other hand more practical papers often give a too descriptive account of DA leaving much to be desired in terms of conceptual consistency.
\end{itemize}
In this pedagogical review our aim is to try to fill in the above gaps. We consider however that the subject of DA is just too broad for an introductory paper to aim at comprehensiveness. This would lead to compile a tedious catalog of methods with little room for readable explanations. Rather, our commitment is to describe 4 important practical cases of DA, emphasizing on consistency. From an intuitive and geometric point of view to start with, then from a more mathematical perspective emphasizing clarity of concepts and, lastly, concluding with a short description of one practical application of the DA method being described.
\\

The 4 special cases of DA we consider are described below. Henceforth, the source domain will refer to the population from which training observations are drawn while the target domain will refer to the population from which test observations are drawn.
\begin{enumerate}
\item \textbf{Prior shift} refers to a situation in which the label distributions are different in the source and target domains. The class conditional distributions of the features given the label are however assumed to be identical in both domains. Such a situation occurs when training observations are selected depending on their label value, for example using a stratified sampling strategy, while ignoring their actual distribution in the test set. Section \ref{prior_shift_section} is devoted to this case.
\item \textbf{Covariate shift} refers to a situation in which the distributions of the features are different but known in the source and the target domains. The conditional dependence of the labels on features are however assumed to be the same in both domains. Such a situation occurs when some observations are more difficult to pick than others resulting in selection bias. This case is described in section \ref{cov_shift_subsection}.
\item \textbf{Concept shift} refers to the case where the dependence of the label upon the features differs between the target and the source domains, often depending on time in which case it is termed a concept drift. The distribution of the features are nevertheless assumed to be the same in both domains. This case is described in section \ref{concept_shift_subsection}.
\item \textbf{Subspace mapping} describes a situation where observations are distributed alike as physical objects in the source and target domains but where the features used to describe them in one or the other are different and related by an unknown change of coordinates. Think for instance of the same object seen under different angles. This situation is dealt with in section \ref{domain_shift_subsection}. 
\end{enumerate}
We examine these 4 instances of DA in turn, using the two main theoretical frameworks of Machine Learning (ML), namely the PAC learning framework of statistical learning and the maximum likelihood principle which is perhaps more familiar to hands-on data scientists. In a nutshell, the PAC theory formulates the aim of ML as making good predictions directly from samples of a probability distribution from which nothing is assumed. The maximum likelihood principle on the other hand assumes that samples are generated from some parametrized probability distribution whose parameter are to be optimized to make the observed samples as likely as possible. Section \ref{stat_learning_recap} is a compact recap of these two point of views of ML that we include for convenience. Readers familiar with these concepts can directly jump to the heart of the matter in section \ref{zoom_on_4_ins_DA}.

\section{Statistical Learning recap}
\label{stat_learning_recap}

\subsection{PAC learning framework}
\label{PAC_learning_subsection}
The classical mathematical framework for defining  statistical learning is called \emph{Probably Approximately Correct} (PAC) learning \cite{UML} for reasons that we shall understand shortly. We will use it in the analysis of the covariate, concept and subspace mappings.
\\

We assume that the observations are defined as pairs $(\bx,y)$ of features $\bx$ in some feature space $\mathcal{X}\subset\R^p$ and of labels $y$ in a label space $\mathcal{Y}$ which could be either a part of $\R$ for a regression or a set of labels for a classification. We assume that the relationship between $\bx$ and $y$ is described by an unknown joint probability distribution\footnote{Both $\mathcal{X}$ and $\mathcal{Y}$ could be either continuous or discrete spaces, but we shall treat this difference informally. Therefore, depending on context, $p(\bx,y)$ could either denote an actual probability or a probability density.} $p(\bx,y)$. Suppose that information is given to us only in the form of a sample $S=\{(\bx_1,y_1),...,(\bx_m,y_m)\}$ of observations drawn from this unknown distribution $p(\bx,y)$. In order to make useful predictions, our aim is to use this information $S$ to find a “good” approximation of the dependence of $y$ on $\bx$ as a function $h:\mathcal{X}\rightarrow\mathcal{Y}$. We select this function from a collection $\mathcal{H}$, the hypothesis class, that we look as plausible candidates for predictors. To assess the quality of a predictor $h$ we assume moreover that we are given a loss function $\ell:\mathcal{Y}\times\mathcal{Y}\rightarrow\R$ which measures the discrepancy $\ell(y,\hat{y})$ between an observed value $y$ and a prediction $\hat{y}=h(\bx)$. Common choices are the $\ell_{0-1}$ loss which counts the number of points where $y\neq y'$ for a binary classifier and $\ell_{\mathrm{LS}}:=|y-y'|^2$ for least square regression. The true risk $R[h]$ associated with a predictor $h\in\mathcal{H}$ is then defined as the expectation of this loss $\ell$ when averaged over the unknown distribution $p$:
\begin{equation}
	\label{true_Risk}
	R[h]:=\E_{(\bx,y)\sim p}[\ell(y,h(\bx))].
\end{equation}
This is the quantity that we want to minimize over predictors $h\in\mathcal{H}$ using some machine learning algorithm. We are thus lead to the following definition :
\paragraph{In intuitive terms:}
a class $\mathcal{H}$ of functions is termed \emph{learnable} when there exists an algorithm $A$ that takes a sample $S$ as input to select a predictor $h_S=A(S)$ from $\mathcal{H}$ which has high chances (“\emph{Probably}”) to be have low risk $R[h]$ (“\emph{Approximately Correct}”), provided the sample size $m:=|S|$ is large enough.
\paragraph{More formally:} a hypothesis class $\mathcal{H}$ is \emph{learnable} if the following is true: there exists an algorithm $A$ which, for any given precision $\epsilon>0$ and $\delta>0$, takes a sample $S$ of size $m$, whose observations are sampled from $p$, and returns a predictor $h_S=A(S)\in\mathcal{H}$ such that
\begin{equation}
	\label{PAC_true_risk}
	R[h_S]\leq\min_{h\in\mathcal{H}}R[h]+\epsilon
\end{equation}
with a confidence $1-\delta$ provided $m:=|S|$ is large enough. In other words the algorithm $A$ should manage to pick a predictor whose risk is  $\epsilon$–close to the optimal $h$ that can be achieved within the class $\mathcal{H}$. If the actual relationship between $\bx$ and $y=h(\bx)$ is deterministic and if moreover this $h$ belongs to $\mathcal{H}$, then the right hand side of (\ref{PAC_true_risk}) simply reduces to $\epsilon$.
\\

PAC theory allows to prove various bounds, either on the minimal size $m$, the so called sample complexity, which guarantees a precision $\epsilon$ with a confidence $1-\delta$ for a given hypothesis class $\mathcal{H}$ or, the other way around, bounds for the precision $\epsilon$ as a function of $m$ and $\delta$. Bounds typically look like
\begin{equation}
	\label{epsilon_PAC_bound}
	\epsilon \leq \mathrm{const}\sqrt{\frac{d[\mathcal{H}]+\log(1/\delta)}{m}},
\end{equation}
where $d[\mathcal{H}]$ measures the complexity of the class $\mathcal{H}$. For a prescribed accuracy $\epsilon$, the higher the complexity $d$, the larger the sample size $m$ should be. There are various notions of complexity, the most famous one being probably the \emph{VC-dimension}\footnote{The VC-dimension $d$ of a set of binary classifiers $\mathcal{H}$ is defined as follows. For a given set ${\bx_1,...,\bx_n}$ of points in $\mathcal{X}$, consider its image through $h\in\mathcal{H}$. This is a set of $n$ binary labels $\{0,1\}$. Now ask for the largest possible size $n$ such that, when $h$ runs over $\mathcal{H}$, we create \emph{all} $2^n$ possible binary assignments. This maximum size is by definition the VC-dimension $d$ of $\mathcal{H}$.}. They all somehow make precise the notion of the richness of a class of functions $\mathcal{H}$. 

\paragraph{Empirical Risk Minimization (ERM) and regularization:} one simple strategy to find an approximate minimizer $h_S$ in (\ref{PAC_true_risk}) is to minimize an empirical version $\hat{R}_S$ of the true risk for a sample $S$ of size $m$:
\begin{equation}
	\label{ERM}
	\hat{R}_S[h]:=\frac{1}{m}\sum_{i=1}^m \ell[y_i,h(\bx_i)].
\end{equation}
This ERM strategy works provided $\hat{R}_S[h]$ is a good enough approximation of $R[h]$ for all $h\in\mathcal{H}$. Fortunately, PAC theory allows to prove bounds like:
\begin{equation}
	\label{EMR_bound}
	R[h]\leq\hat{R}_S[h] + \mathrm{const}\sqrt{\frac{d[\mathcal{H}]+\log(1/\delta)}{m}}\textnormal{  for all  }h\in\mathcal{H},
\end{equation} 
with a confidence $1-\delta$ when observations in $S$ are sampled from $p$. If data is not scarce, just take $m=|S|$ so large that is makes the second term on the right of (\ref{EMR_bound}) as small as we wish, and define the ERM algorithm as $A_{\mathrm{ERM}}(S):=\argmin_{h\in\mathcal{H}}\hat{R}_S[h]$. In practice though, the amount of data is often limited. Therefore we could try to minimize the true risk $R$ by minimizing this time the sum of both term on the right hand side of (\ref{EMR_bound}). This implies solving a bias-complexity tradeoff because on one hand, a rich class $\mathcal{H}$ will allow to find a low value for the first term $\hat{R}_S[h]$ in (\ref{EMR_bound}) but on the other, we must pay for this with a high complexity $d[\mathcal{H}]$ in the second term.
\\

PAC theory thus provides us with a very clean conceptual framework for statistical learning. In particular it does not assume any prior knowledge regarding the unknown distribution $p$ and uses solely the data $S$ at hand.

\subsection{Maximum likelihood framework}
The maximum likelihood approach (MLA) to machine learning is probably the best known for data scientists. We shall use it in our treatment of the prior shift. Therefore we quickly recall it here and contrast it with the PAC framework. 
\\

The MLA tries to learn the unknown distribution $p(\bx,y)$ of the data. We say therefore that it looks for a \emph{generative model} of the data that will be used to make predictions. This is in marked contrast with the PAC framework which bypasses this step and focuses on directly finding a good predictor $h$, looking thus for a \emph{discriminant model}. Learning $p$ is usually harder because from its knowledge we can, at least in principle, deduce a predictor $h$ but not the other way around. For instance a binary classifier which minimizes the $\ell_{0-1}$ loss is given by $h_{\mathrm{Bayes}}(\bx):=\argmax_{y\in\{0,1\}}p(y|\bx)$. The MLA on the other hand allows including prior knowledge on $p$ by restricting the search over $p$ to a set of parametrized densities $p_\theta(\bx,y)$. The MLA stipulates that a reasonable choice for the parameters $\theta$ is the one which maximizes the (log) likelihood $L_\theta(S)$ of a sample $S=\{(\bx_1,y_1),...(\bx_m,y_m)\}$
\begin{equation}
	\label{LogLikelihood}
	L_\theta(S):=\log \left(\prod_{i=1}^m p_\theta(\bx_i,y_i)\right).
\end{equation}
Maximizing (\ref{LogLikelihood}) with respect to $\theta$ can be cast as an ERM problem for a specific loss, namely for the log-loss $\ell_\theta(\bx,y):=-\log  p_\theta(\bx,y)$. The maximum likelihood estimator $\theta_{\mathrm{ML}}$ for the parameter $\theta$ is thus
\begin{equation}
	\theta_{\mathrm{ML}}:=\argmax_\theta L_\theta(S) = \argmin_\theta \sum_{i=1}^m \ell_\theta(\bx_i,y_i).
\end{equation}
For a classification problem $\mathcal{Y}=\{\omega_1,...,\omega_K\}$ is a finite set of labels. Let $z_{ik}=1$ when $y_i=\omega_k$ and $0$ otherwise. The negative log likelihood is often written as
\begin{equation}
	\label{NLL}
	-L_\theta(S)=-\sum_{i=1}^m \sum_{k=1}^K\: z_{ik}\log p_\theta(\bx_i,\omega_k)
\end{equation}
The advantage of the MLA to statistical learning is that it allows to easily include prior knowledge about the true distribution $p$. The caveat regarding MLA is that it usually suffers from over-fitting and (\ref{NLL}) should thus be regularized.

\section{Zoom on four instances of domain adaptation}
\label{zoom_on_4_ins_DA}
In the following four subsections we examine in turn four cases of domain adaptations which are of importance in practice. The 4 subsections can be read independently. From here on we shall use an ‘S' subscript in $p_\mathrm{S}$ to denote the \emph{source} distribution from which the training set $S$ is sampled and a ‘T' subscript in $p_\mathrm{T}$ to denote the distribution of the \emph{target} observations to which we intend to apply our predicting model. We systematically use hats to denote estimators of the various quantities of interest, like $\hat{R}_\mathrm{S}[.]$ or $\hat{p}_\mathrm{T}(.)$.

\subsection{Prior shift}
\label{prior_shift_section}

\subsubsection{Intuition}
\begin{figure}[h]
\centering
\includegraphics[scale=0.25]{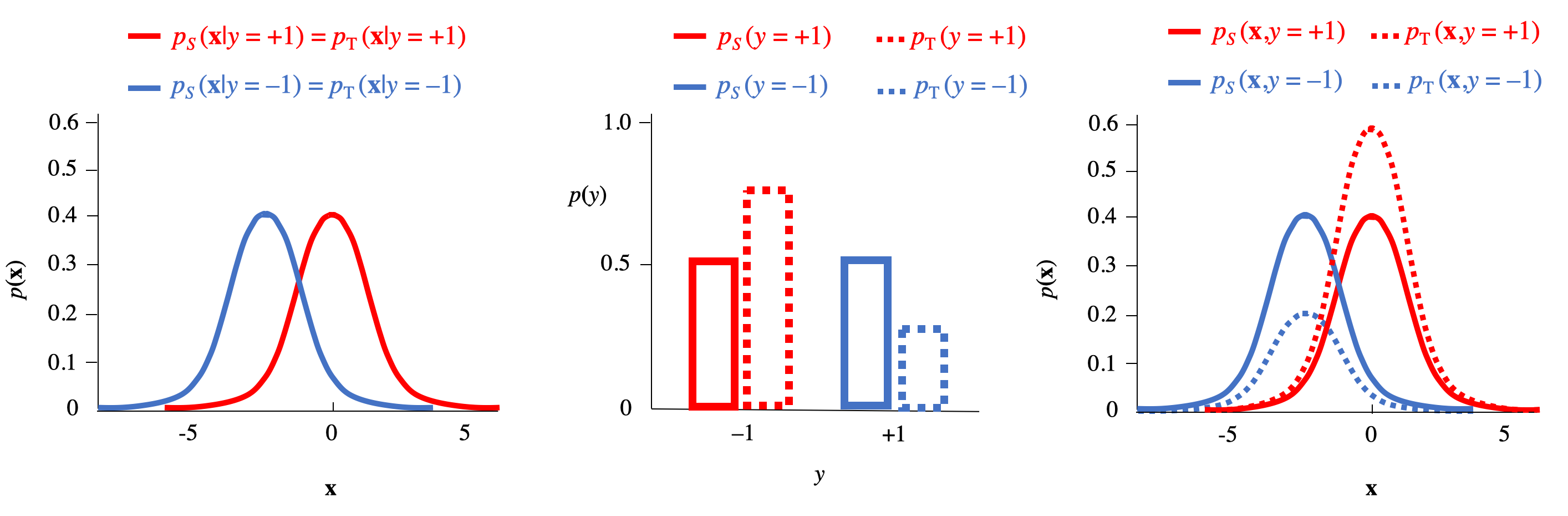}
\caption{{\small The prior shift corresponds to a situation in which the prior distributions $p_\mathrm{S}(y)$ and $p_\mathrm{T}(y)$ on labels are different in the source and in the target (middle). The class conditional distributions $p_\mathrm{S}(\bx|y)=p_\mathrm{T}(\bx|y)$ on the other hand are supposed to be the same in both domains (left). The resulting joint distributions are shown on the right.}}
\label{prior_shift_fig} 
\end{figure}

Prior shift refers to a situation in which the source distribution $p_\mathrm{S}$ used for picking the training observations is biased with respect to the target distribution $p_\mathrm{T}$ because the priori distribution of the labels $y_i$ in both domain are different. We will focus here on classification where $\mathcal{Y}=\{\omega_1,...,\omega_K\}$ is a finite set of labels. The most interesting and difficult case is one in which the prior distribution of the labels in the target is unknown while observations in the source are selected according to their label using some known strategy. Stratified sampling for instance selects an equal number of observations $(y_j=\omega_k, \bx_j)$ in each class $\omega_k$. The class conditional distributions $p_\mathrm{S}(\bx|y)=p_\mathrm{T}(\bx|y)$ on the other hand are supposed to be the same in both domains. Figure \ref{prior_shift_fig} illustrates the situation.
\subsubsection{What mathematics tells us}
The basic assumptions for a prior shift are that the prior distributions on labels are different $p_\mathrm{S}(y)\neq p_\mathrm{T}(y)$ but that the distributions for the class conditional feature distributions are equal $p_\mathrm{S}(\bx|y) = p_\mathrm{T}(\bx|y)$. If we were fortunate enough to know the target prior distribution on the labels we could try to correct the empirical risk (\ref{ERM}), defined as an expectation with respect to $\hat{p}_\mathrm{S}$, with a factor $\hat{p}_\mathrm{T}(y)/\hat{p}_\mathrm{S}(y)$. We shall indeed use such a strategy in section \ref{cov_shift_subsection} when we examine the covariate shift (see (\ref{R_T_cov_shift_corrected}) and (\ref{R_T_cov_shift_emp})). However, in this subsection we shall focus on the harder and more realistic situation in which $p_\mathrm{T}(y)$ is unknown to us. For this reason, following \cite{adust_outputs}, we now switch to the MLA (see section \ref{stat_learning_recap} for notations). 
\\

Our aim is to find a good approximation for the target conditional distribution $p_\mathrm{T}(y|\bx)$ from which a classifier $h(\bx)$ can later be defined. We assume however that we have used our biased data source to draw a sample $S=\{(\bx_1,y_1),...(\bx_m,y_m)\}$ to train an algorithm and obtain a probabilistic model\footnote{Logistic regressions or neural network with softmax output layer are examples of such predictors.} $\hat{p}_\mathrm{S}(y|\bx)$ for the true posterior $p_\mathrm{S}(y|\bx)$. Our strategy is to compute what we want, a good approximation $\hat{p}_\mathrm{T}(y|\bx)$ of the posterior in the target, using what we have, namely the posterior in the source $\hat{p}_\mathrm{S}(y|\bx)$ and the prior in the source $\hat{p}_\mathrm{S}(y=\omega_k)$ which is readily computed as the proportion $m_k/|S|$ of training observations whose label equals $\omega_k$.
\\

First, starting from the equality assumption between the (empirical) class conditionals $\hat{p}_\mathrm{S}(\bx|y) = \hat{p}_\mathrm{T}(\bx|y)$, using two relations 
\[
	\hat{p}_\mathrm{D}(\bx|y)=\frac{\hat{p}_\mathrm{D}(y|\bx)\hat{p}_\mathrm{D}(\bx)}{\hat{p}_\mathrm{D}(y)}\textnormal{ for D=‘S’ or D=‘T’,}
\]
solving for the posterior $\hat{p}_\mathrm{T}(\omega_k|\bx):=\hat{p}_\mathrm{T}(y=\omega_k|\bx)$ and using the fact that $\hat{p}_\mathrm{T}(\omega_k|\bx)$ should sum to $1$ we readily get
\begin{equation}
	\label{corrected_posterior}
	\hat{p}_\mathrm{T}(\omega_k|\bx)=\frac{\hat{w}(\omega_k)\,\hat{p}_\mathrm{S}(\omega_k|\bx)}{\displaystyle\sum_{k'=1}^K \hat{w}(\omega_{k'})\,\hat{p}_\mathrm{S}(\omega_{k'}|\bx)}
	\textnormal{ where } \hat{w}(\omega_k):=\frac{\hat{p}_\mathrm{T}(\omega_{k})}{\hat{p}_\mathrm{S}(\omega_k)}. 
\end{equation}
The interpretation of (\ref{corrected_posterior}) is straightforward: to get the desired posterior in the target domain, we simply correct the source posterior $p_\mathrm{S}(\omega_k|\bx)$ with the reweighing ratio\footnote{Notice that the reweighing factor $\hat{w}(\omega_k)$ is an estimate of the ratio we mentioned earlier and that would use to correct the risk in the PAC framework, if we knew the target prior $p_\mathrm{T}(y)$.} $\hat{p}_\mathrm{T}(\omega_k)/\hat{p}_\mathrm{S}(\omega_k)$ which corrects for the bias in the source label distribution. But wait, $\hat{p}_\mathrm{T}(\omega_k)$ is unknown! It thus looks like arriving at (\ref{corrected_posterior}) we have been going in circles. Well, not really. To see this, let $\bx'_1,...,\bx'_m$ be the features of a  \emph{new} set of observations drawn from the \emph{target} domain and for which we would like to compute the posterior $\hat{p}_\mathrm{T}(\omega_k|\bx'_i)$. Let's now rewrite (\ref{corrected_posterior}) as a fixed point equation for the unknown values $\hat{p}_\mathrm{T}(\omega_k)$ we are interested in:
\begin{equation}
	\label{prior_fixed_point}
	\begin{split}
		\hat{p}_\mathrm{T}(\omega_k) &= 
        \sum_{i=1}^m \hat{p}_\mathrm{T}(\omega_k|\bx'_i)\hat{p}_\mathrm{T}(\bx'_i)
        =\frac{1}{m}\sum_{i=1}^m \hat{p}_\mathrm{T}(\omega_k|\bx'_i),\\	
		\textnormal{ where }\hat{p}_\mathrm{T}(\omega_k|\bx'_i) &= \frac{\displaystyle{\frac{\hat{p}_\mathrm{T}(\omega_k)}{\hat{p}_\mathrm{S}(\omega_k)}}  \,\hat{p}_\mathrm{S}(\omega_k|\bx'_i)}{\displaystyle\sum_{k'=1}^K \frac{\hat{p}_\mathrm{T}(\omega_{k'})}{\hat{p}_\mathrm{S}(\omega_{k'})}\,\hat{p}_\mathrm{S}(\omega_{k'}|\bx'_i)}.
        \end{split}
\end{equation}
On intuitive grounds we assumed that we could set $\hat{p}_\mathrm{T}(\bx'_i)=1/m$ in the first line of (\ref{prior_fixed_point}). This will be justified rigorously in the Appendix \ref{appendix:a}.  Remember that the source prior $\hat{p}_\mathrm{S}(\omega)$ and the source posterior $\hat{p}_\mathrm{S}(\omega|\bx)$ in (\ref{prior_fixed_point}) are all known. Fixed point equations are nice because they can be solved by iteration. Let's thus transform  (\ref{prior_fixed_point}) into an iteration that generates a sequence of approximations $\hat{p}^{(s)}_\mathrm{T}(\omega_k|\bx'_i)$ and $\hat{p}^{(s)}_\mathrm{T}(\omega_k)$ which converges, hopefully, towards good approximations $\hat{p}_\mathrm{T}(\omega_k|\bx'_i)$ and $\hat{p}_\mathrm{T}(\omega_k)$ of the true target distributions:
\begin{equation}
	\label{prior_fixed_point_iter}
	\begin{split}
	\hat{p}^{(s+1)}_\mathrm{T}(\omega_k) &= \frac{1}{m}\sum_{i=1}^m \hat{p}^{(s)}_\mathrm{T}(\omega_k|\bx'_i),\\
		\textnormal{ where }\hat{p}^{(s)}_\mathrm{T}(\omega_k|\bx'_i) &= \frac{\displaystyle{\frac{\hat{p}^{(s)}_\mathrm{T}(\omega_k)}{\hat{p}_\mathrm{S}(\omega_k)}}  \,\hat{p}_\mathrm{S}(\omega_k|\bx'_i)}{\displaystyle\sum_{k'=1}^K \frac{\hat{p}^{(s)}_\mathrm{T}(\omega_{k'})}{\hat{p}_\mathrm{S}(\omega_{k'})}\,\hat{p}_\mathrm{S}(\omega_{k'}|\bx'_i)},
        \end{split}
\end{equation}
for $s\geq 0$ and using $\hat{p}^{(0)}_\mathrm{T}(\omega):=\hat{p}_\mathrm{S}(\omega)$ as an initial approximation of the unknown target prior. Convergences certainly has to be checked in practice and could depend on a reasonable choice of the initial estimate $\hat{p}^{(0)}_\mathrm{T}(\omega)$ of the true prior. 
\\

Interestingly, the iteration procedure (\ref{prior_fixed_point_iter}) can be seen as be an instance of the classical  EM algorithm for finding the maximum likelihood for a distribution where the latent variables are the unobserved classes $\omega_k$ and the parameters are their unknown prior probabilities $p_\mathrm{T}(\omega_k)$. We refer to Appendix \ref{appendix:a} for a proof of this assertion.

\paragraph{A statistical test}
Experience has shown that computing the posterior probabilities  using the iterative method (\ref{prior_fixed_point_iter}) can actually worsen the error rate, as compared to doing no corrections, when such the source and target priors are not significantly different \cite{adust_outputs}. Indeed, even if the true distributions $p_\mathrm{S}$ and $p_\mathrm{T}$ are strictly equal, we could be so unlucky to pick samples $S$ and $S'$ for which our iterative procedure (\ref{prior_fixed_point_iter}) converges to a $\hat{p}_\mathrm{T}$ which is significantly different from $\hat{p}_\mathrm{S}$.
\\

Fortunately, a simple statistical test can help us. It simply involves computing the ratio of the corrected target likelihood to the uncorrected source likelihood of the observations $\{\bx_1,...,\bx_m\}$. For any given class $\omega_k$ and for D=‘S’ or D=‘T’ we have:
\begin{equation}
\begin{split}
	L_\mathrm{D}(\bx_1,...,\bx_m) 
	&:= \prod_{i=1}^m \hat{p}_\mathrm{D}(\bx_i)\\
	&=  \prod_{i=1}^m \left(\frac{\hat{p}_\mathrm{D}(\bx_i|\omega_k)\,\hat{p}_\mathrm{D}(\omega_k}{\hat{p}_\mathrm{D}(\omega_k|\bx_i)}\right).
\end{split}
\end{equation}
Using the hypothesis that the class conditionals are equal $\hat{p}_\mathrm{S}(\bx_i|\omega_k)=\hat{p}_\mathrm{T}(\bx_i|\omega_k)$ we obtain, for any $\omega_k$:
\begin{equation}
	\log\left(\frac{L_\mathrm{T}(\bx_1,...,\bx_m) }{L_\mathrm{S}(\bx_1,...,\bx_m) }\right)
	= \sum_{i=1}^m \log\left(\frac{\hat{p}_\mathrm{S}(\omega_k|\bx_i)}
	                                                {\hat{p}_\mathrm{T}(\omega_k|\bx_i)}\right)
	+ m \log\left(\frac{\hat{p}_\mathrm{T}(\omega_k)}
	                             {\hat{p}_\mathrm{S}(\omega_k)}\right),
\end{equation}
which can be computed once the iteration procedure (\ref{prior_fixed_point_iter}) has converged to $\hat{p}_\mathrm{T}(\omega_k|\bx_i)$ and $\hat{p}_\mathrm{T}(\omega_k)$. These quantities can be shown to be asymptotically distributed as a $\chi^2_{(K-1)}$ for large samples, $K$ being the number of classes $\omega_k$ \cite{Hook_Craig}. If this log likelihood ratio has low probability according to this law, we conclude that the source and target distributions are significantly different and that the iterative correction procedure for the prior and posterior target distributions is worthwhile. 

\subsubsection{In practice}
The EM method discussed above for computing the prior a posterior target distribution was tested on three real world medical data sets in \cite{adust_outputs}. All were binary classification problems\footnote{The three data sets were: Pima Indian Diabetes, Breast Cancer Wisconsin and Bupa Liver Disorders.}. Each data set contained a few hundreds examples from which balanced training data set were created: $p_\mathrm{S}(\omega_1)=p_\mathrm{S}(\omega_2)=0.5$. The remaining data set was then used to create artificially biased test data with $p_\mathrm{T}(\omega_1)=0.20$ and $p_\mathrm{T}(\omega_2)=0.80$. The classifier used here is a multilayer neural network. The experiment was replicated with ten different splits of the data into training and test sets, and for each such split the training was repeated ten times. The average results for these 100 experiments is shown in table \ref{prior_shift_experiment} below.

\begin{table}[h!]
\centering
\begin{tabular}{|c|c|c|c|c|c|}
\hline Data  & True      & Priors estimated  &  \multicolumn{3}{|c|}{\% of correct classification}\\  \cline{4-6} 
     set                   & priors    & by EM                  &  No adjustment           & EM           & True priors \\
\hline 
\hline
Diabetes & 20\% & 24.8\% & 67.4\% & 76.3\% & 78.\% \\ 
\hline 
Breast     & 20\% & 18.0\% & 91.3\% & 92.0\% & 92.6\% \\ 
\hline 
Liver       & 20\% & 24.6\% & 68.0\% & 75.7\% & 79.1\% \\ 
\hline 
\end{tabular} 
\caption{\small Efficiency of the EM method on three real medical data sets used in\cite{adust_outputs}.}
\label{prior_shift_experiment}
\end{table}
In summary the experiments in \cite{adust_outputs} showed that the EM method using only a small number of iterations ($\leq 5$) provides both good estimates of the true priors and a classification accuracy close to what the true priors would yield.

\subsection{Covariate shift}
\label{cov_shift_subsection}

\subsubsection{Intuition}
\begin{figure}[h]
\centering
\includegraphics[scale=0.27]{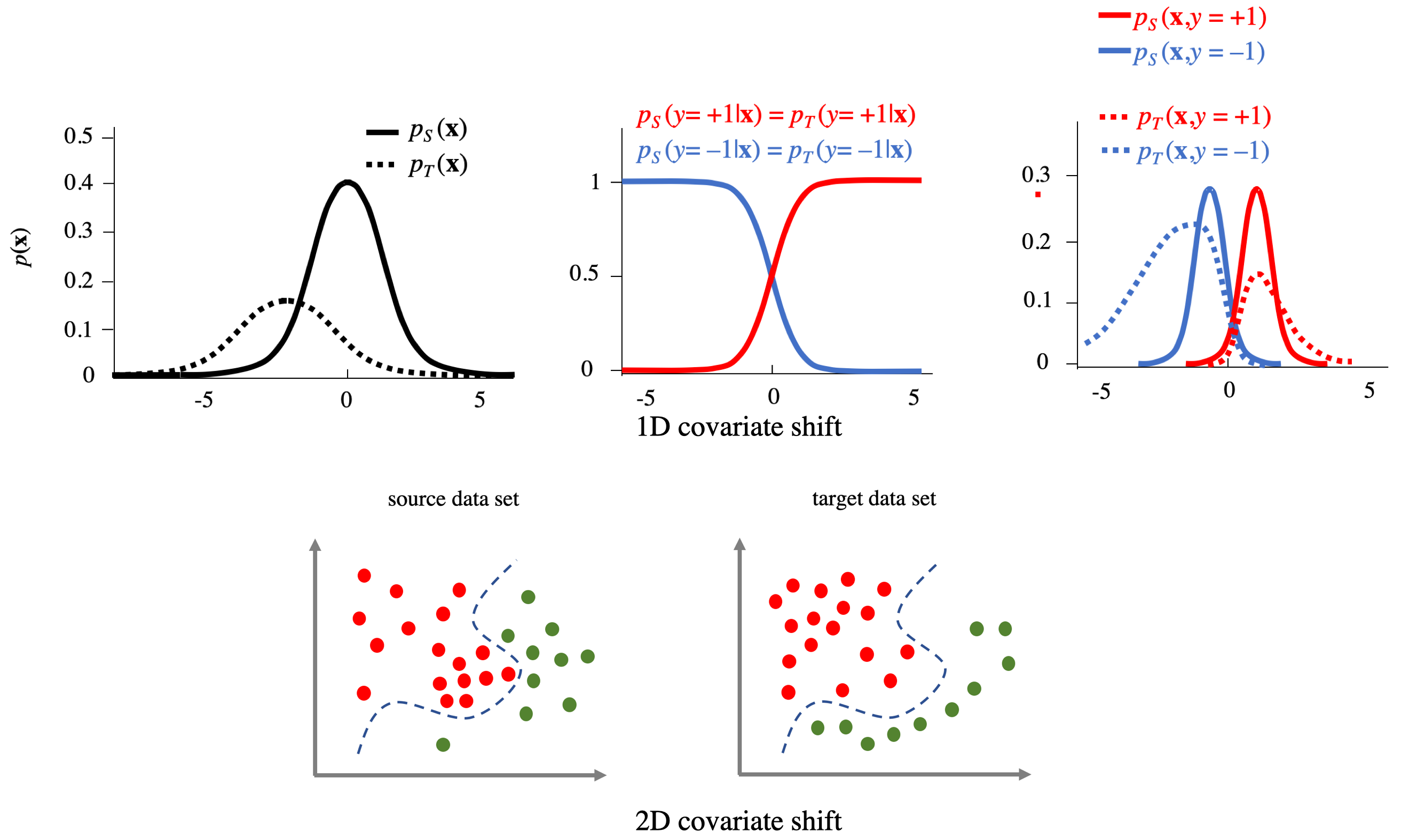}
\caption{{\small The covariate shift corresponds to a situation where the source $p_{\mathrm{S}}(\bx)$ and target $p_{\mathrm{T}}(\bx)$ features distribution are different. Top: a 1D example. Bottom: a 2D example where the dashed curve symbolically represents the unchanged conditional probability $p_{\mathrm{S}}(y|\bx)=p_{\mathrm{T}}(y|\bx)$.}}
\label{cov_shift_fig} 
\end{figure}

Covariate shift\footnote{Also sometimes termed “real concept drift” like in \cite{A_survey_on_CDA}.} describes a situation where the source distribution is biased because the training observations were sampled depending on their features $\bx$ in proportions $p_{\mathrm{S}}(\bx)$ that do not match the target distribution $p_{\mathrm{T}}(\bx)$. This should be contrasted with the prior shift we discussed in section \ref{prior_shift_section} where a bias existed in the source distribution because observations were selected according to their label $y$ in proportions $p_{\mathrm{S}}(y)$ that did not match $p_{\mathrm{T}}(y)$. Covariate shift assumes however that the dependency of the response $y$ on the features $\bx$, as described by the conditional probability $p_{\mathrm{S}}(y|\bx)$ in the source, is the same as in the  target population $p_{\mathrm{T}}(y|\bx)$. The situation is depicted in figure \ref{cov_shift_fig} for the cases where $\bx$ is either 1D or 2D. 
We emphasize that the target feature distribution $p_{\mathrm{T}}(\bx)$ is assumed to be known in the presence of a covariate shift. This  again is in contrast with the priori shift where the target label distribution $p_{\mathrm{T}}(y)$ was assumed to be unknown. In other words, covariate and prior shift are \textit{not} symmetric cases that would result by simply swapping features $\bx$ with labels $y$.
\\

A covariate shift typically occurs when the cost or the difficulty of picking an observation with given features $\bx$ strongly impacts the probability of selecting an observation $(\bx,y)$ thus making it practically impossible to replicate the target feature distribution $p_{\mathrm{T}}(\bx)$ in the training set. A typical example could be a survey were some categories of individuals are more difficult to reach than others while the model should be able to classify all categories equally well.

\subsubsection{What mathematics tells us}
In a covariate shift the source and target distributions on observations differ only because their respective marginal distribution on features are distinct $p_{\mathrm{S}}(\bx)\neq p_{\mathrm{T}}(\bx)$. The predictive distribution on labels given the features on the other hand is the same in both domains: $p_{\mathrm{S}}(y|\bx) = p_{\mathrm{T}}(y|\bx)$. 
The main question then is: “Can we somehow compensate for using the wrong training distribution?” The short answer is: “Yes, we can adjust the training procedure but only provided the marginal distributions on features in the source and the target are not too far apart in a sense we will make precise soon.” Notice that this is once more in contrast with the prior shift case where compensation happens only \emph{after} training, when using the EM–procedure on the posterior distribution. Let's now look at this in more detail. 

Covariate shift is by far the case of domain adaptation that has been most studied from a theoretical perspective, especially within the PAC framework that we will also follow here. Recall from section \ref{PAC_learning_subsection} that the aim is to find a good predictor $h(\bx)$, namely one which has low expectation for the true (target) risk $R_{\mathrm{T}}[h]$ 
\begin{equation}
	\label{R_T_cov_shift}
	R_{\mathrm{T}}[h]:=\E_{p_{\mathrm{T}}}[\ell(h(\bx),y)]
\end{equation}
As we only have access to samples from $p_{\mathrm{S}}(\bx,y)$ we rewrite the above as 
\begin{equation}
\label{R_T_cov_shift_corrected}
\begin{split}
	R_{\mathrm{T}}[h] &= \E_{p_{\mathrm{S}}}\left[\frac{p_{\mathrm{T}}(\bx,y)}{p_{\mathrm{S}}(\bx,y)}\:\ell(h(\bx),y)\right] 
                   = \E_{p_{\mathrm{S}}}\left[\frac{p_{\mathrm{T}}(\bx)}{p_{\mathrm{S}}(\bx)}\:\ell(h(\bx),y)\right] \\
                 &= \E_{p_{\mathrm{S}}}\left[w(\bx)\:\ell(h(\bx),y)\right]
\end{split}
\end{equation}
where we define the re-weighting factor $w(\bx):=p_{\mathrm{T}}(\bx)/p_{\mathrm{S}}(\bx)$. If we had a good approximation $\hat{w}(\bx)$ for the re-weighting factor we could use if to build \linebreak a corrected version $R_{S,w}$ of the empirical risk for a sample $S=$ \linebreak  $\left\lbrace(\bx_1,y_1),...,(\bx_m,y_m) \right\rbrace$ drawn from $p_{\mathrm{S}}(\bx,y)$:
\begin{equation}
	\label{R_T_cov_shift_emp}
	\hat{R}_{S,w}[h] := \frac{1}{m} \sum_{i=1}^m w(\bx)\:\ell(h(\bx_i),y_i)
\end{equation}
If the size $m$ of the sample $S$ is large enough we can hope that $\hat{R}_{S,w}[h]$ will be a good approximation for $R_{\mathrm{T}}[h]$ and therefore that the minimizer $h_{\mathrm{S}}$ of the former will be a decent approximation for the minimizer of the latter. As a matter of fact, a rigorous analysis that extends the basic PAC theory confirms this \cite{Learning_bounds_for_IW}. Indeed one can prove that when selecting the sample $S$ according to $p_{\mathrm{S}}(\bx,y)$, the following bound holds with a confidence $1-\delta$:
\begin{equation}
	\label{cov_shift_bound}
	R_{\mathrm{T}}[h] \leq \hat{R}_{S,w}[h] + C\sqrt{1+\mathrm{Var}_{p_{\mathrm{S}}}[w]}
	\sqrt[\leftroot{-3}\uproot{6}\frac{3}{8}]{\frac{\tilde{d}(\cal{H})}{m}\log \left(2e\frac{m}			
	{\tilde{d}(\cal{H})} \right)+\frac{\log\frac{4}{\delta}}{m}}
\end{equation}
for any $h$ belonging to the hypothesis class $\mathcal{H}$. Let's quickly comment on this formula. It clearly shows the role of the ratio of the sample size $m$ to the complexity $\tilde{d}(\cal{H})$ which is a variant\footnote{It is called pseudo-dimension of the class $\cal{H}$ \cite{Learning_bounds_for_IW}} of the complexity $d(\cal{H})$ in (\ref{epsilon_PAC_bound}). The decay $m^{-\frac{3}{8}}$ with the sample size is slightly slower than what the standard PAC bound (\ref{EMR_bound}) predicts. The weight $w$ that enters in (\ref{cov_shift_bound}) is the true ratio of the marginals $p_{\mathrm{T}}(\bx)/p_{\mathrm{S}}(\bx)$ (that we actually do not know in practice).
The most significant element in (\ref{cov_shift_bound}) is the $\mathrm{Var}_{p_{\mathrm{S}}}[w]:=\E_{p_{\mathrm{S}}}[w^2]-\left(\E_{p_{\mathrm{S}}}[w]\right)^2$ term which denotes the variance of the true reweighting factor $w$ under $p_{\mathrm{S}}(\bx)$. This allows us to understand under what circumstances we can to hope correct the covariate shift with the re-weighting factor $w$. 

\paragraph*{When can we fix the covariate shift?}
When the source $p_{\mathrm{S}}$ and the target $p_{\mathrm{T}}$ are very different, the ratio $w$ will fluctuate and thus its variance will penalizes the accuracy of the empirical estimation of $R_{\mathrm{T}}$ by $\hat{R}_{S,w}$ in (\ref{cov_shift_bound}). Notice in particular that for $w$ to remain finite, $p_{\mathrm{S}}(\bx)$ is not allowed to vanish in the support of the target distribution $p_{\mathrm{T}}(\bx)$. This certainly corresponds to our intuition that $p_{\mathrm{S}}$ should roughly sample the same points as $p_{\mathrm{T}}$. To get a better intuition for $\mathrm{Var}_{p_{\mathrm{S}}}[w]$ we can evaluate this quantity for two normal distributions, say $p_{\mathrm{S}}\sim\cal{N}(\mu_{\mathrm{S}},\sigma_{\mathrm{S}})$ and $p_{\mathrm{T}}\sim\cal{N}(\mu_{\mathrm{T}},\sigma_{\mathrm{T}})$. An easy calculation shows that $\mathrm{Var}_{p_{\mathrm{S}}}[w]$ remains bounded as long as $\sigma_{\mathrm{S}}>\frac{1}{\sqrt{2}}\sigma_{\mathrm{T}}$ which means that $p_{\mathrm{S}}$ should sample $\bx$ broadly enough with respect to $p_{\mathrm{T}}$.

\paragraph*{How can we fix the covariate shift?} Formula (\ref{cov_shift_bound}) confirms that looking for an approximation $\hat{w}$ of the ratio $w$ in order to minimize the reweighted risk $\hat{R}_{S,w}[h]$ may be worth the effort. The most obvious approach proceeds by first looking for approximations $\hat{p}_{\mathrm{S}}(\bx)$ and $\hat{p}_{\mathrm{T}}(\bx)$ before computing the ratio $\hat{w}:=\hat{p}_{\mathrm{T}}(\bx)/\hat{p}_{\mathrm{S}}(\bx)$. Such approximation are usually found by \emph{Kernel Density Estimation} (KDE) which amounts to putting an appropriate kernel on top of each sample point in the empirical distributions $\hat{p}_\mathrm{S}$ and $\hat{p}_\mathrm{T}$. However, experience shows that computing a quotient of such approximations is a risky business because small errors in the denominator can lead to uncontrolled errors in the estimate $\hat{w}$ of the ratio.

It turns out that a better strategy is to estimate $\hat{w}(\bx)$ directly. For this purpose we reformulate our problem as finding a weight $\hat{w}(\bx)$ that makes the corrected source distribution $\hat{w}(\bx)\:p_{\mathrm{S}}(\bx)$ is in some sense “close” to the target distribution $p_{\mathrm{T}}(\bx)$. Now there are various concepts available for measuring the discrepancy between two probability distributions that we could use for our purpose, like for instance:
\begin{itemize}
\item the expected square error $\E_{p_{\mathrm{S}}}\left[\left(\hat{w}-{p_{\mathrm{T}}/p_{\mathrm{S}}} \right)^2\right]$,
\item the KL divergence $\mathrm{KL}\left(p_{\mathrm{T}}\Vert\hat{w}\:p_{\mathrm{S}}\right):=\E_{p_{\mathrm{T}}}\left[\log\left(p_{\mathrm{T}}/\hat{w}p_{\mathrm{S}}\right)\right]$,
\item the \emph{Maximum Mean Discrepancy}  $\mathrm{MMD}[p_{\mathrm{T}},\hat{w}\:p_{\mathrm{S}}]$ is still another possibility that has become quite popular lately. This is the one we are going to describe in some detail below and that we shall apply to a practical example of covariate shift in section \ref{cov_shift_in_practice}.
\end{itemize}
Minimizing $\mathrm{MMD}[p_{\mathrm{T}},\hat{w}p_{\mathrm{S}}]$ with respect to $\hat{w}$ is known as the \emph{Kernel Mean Matching} (KMM) method. It uses a \emph{kernel trick} similar to the one used for constructing SVM classifiers \cite{Bishop}.

\paragraph*{The KMM method} A Hilbert $H$ space is a great place to work because it is naturally equipped with a metric. It is thus tempting  to map each feature vector $\bx\in\mathcal{X}$ to a vector $\Phi_{\bx}\in H$ because this mapping then induces a proximity notion on $\mathcal{X}$ through the scalar product in $H$, namely $K(\bx,\by):=\langle\Phi_{\bf x},\Phi_{\bf y}\rangle$. The function $K$ defines a so called \emph{kernel function} on the original feature space $\mathcal{X}$. The well known kernel trick uses the observation that, as long as a given kernel $K(\bx,\by)$ satisfies a simple positivity constraint\footnote{Mercer theorem requires that $K$ should be positive definite: $\sum_{i=1}^n \sum_{j=1}^n K(\bx_i,\by_i)\:c_i c_j>0$ for all sequences of vectors $\bx_1,...,\bx_n$ in $\mathcal{X}$ and of sequences of real numbers $c_1,...,c_n$.}, it is possible to find a mapping $\bx\rightarrow\Phi_{\bx}$ such that the given $K(\bx,\by)$ can be represented by a scalar product $\langle\Phi_{\bf x},\Phi_{\bf y}\rangle$. In other words, given an appropriate $K$ we can in principle reconstruct the mapping $\Phi$. One then shows that $\Phi_{\bx}(\by):=K(\bx,\by)$ and that $f(\bx)=\langle\Phi_{\bx},f\rangle$ for any function $f\in H$, which explains why such a Hilbert space $H$ is termed a \emph{Reproducing Kernel Hilbert Space} (RKHS).  

Remember that our aim is to measure the discrepancy between two probability distributions, say $\alpha$ and $\beta$, on $\mathcal{X}$. As $\Phi$ maps features $\bx$ in $\mathcal{X}$ to a vector $\Phi_{\bx}$ in $H$ we can just as well associate such a vector to a distribution $\alpha$ on $\mathcal{X}$ by computing the corresponding expectation $\E_{\bx\sim \alpha}[\Phi_{\bx}]$. The MMD-distance is then naturally defined as:
\begin{equation}
	\label{MMD_def}
	\begin{split}
			\mathrm{MMD}[\alpha,\beta]&:=\Vert\E_{\bx\sim \alpha}[\Phi_{\bx}]-\E_{\bx\sim \beta}
			[\Phi_{\bx}]\Vert \\
        &= \sup_{\Vert f\Vert\leq 1}\langle f,\E_{\alpha}[\Phi_{\bx}]-\E_{\beta}[\Phi_{\bx}] \rangle \\
        &= \sup_{\Vert f\Vert\leq 1}\left(\E_\alpha\langle f,\Phi_{\bx}\rangle]-
             \E_\beta[\langle f,\Phi_{\bx}\rangle]\right) \\
        &=  \sup_{\Vert f\Vert\leq 1}\left(\E_\alpha{[f(}\bx)]-
             \E_\beta[f(\bx)]\right)
	\end{split}
\end{equation}
where we successively used the identity $\Vert g\Vert=\sup_{\Vert f\Vert\leq 1}\langle f,g\rangle$, swapped the expectation $\E[.]$ and the scalar product $\langle.,.\rangle$ and applied the RKHS identity $f(\bx)=\langle\Phi_{\bx},f\rangle$. The last line in (\ref{MMD_def}) shows the the MMD distance between $\alpha$ and $\beta$ is the maximum of the difference between their mean when $f$ is chosen from the unit ball in $H$ , hence the name \emph{Maximum Mean Discrepancy}. For our purpose it will be more convenient to consider the square of the MMD:
\begin{equation}
	\label{MMD_square}
	\left(\mathrm{MMD}[\alpha,\beta]\right)^2=
	\Vert\E_{\alpha}[\Phi_{\bx}]\Vert^2
	-2\langle \E_{\alpha}[\Phi_{\bx}],\E_{\beta}[\Phi_{\bx}]\rangle - 
	\Vert\E_{\beta}[\Phi_{\bx}]\Vert^2
\end{equation}
Let's now plug $\alpha=\hat{w}\hat{p}_{\mathrm{S}}$, the reweighted empirical source distribution, and $\beta= \hat{p}_{\mathrm{T}} $, the empirical target distribution, into (\ref{MMD_square}). The expectation $\E_{\hat{w}p_{\mathrm{S}}}$ becomes the weighted arithmetic mean over the source sample $\bx_1,...,\bx_{m_{\mathrm{S}}}$ and similarly the expectation $\E_{p_{\mathrm{T}}}$ becomes the arithmetic mean over the target sample $\bx'_1,...,\bx'_{m_{\mathrm{T}}}$. We obtain:
\begin{equation}
	\label{MMD_empirical}
	\begin{split}
	\left(\mathrm{MMD}[\hat{w}\hat{p}_{\mathrm{S}},\hat{p}_{\mathrm{T}}]\right)^2 &= \\
	&{ }\;\;\;\;\,\frac{1}{m_{\mathrm{S}}^2}\sum_{i,j=1}^{m_{\mathrm{S}}}\hat{w}(\bx_i)\hat{w}(\bx_j)\:\langle\Phi_{\bx_i},\Phi_{\bx_j}\rangle
\\	
    &{ } 
	- \frac{2}{m_{\mathrm{S}} m_{\mathrm{T}}}\sum_{i=1}^{m_{\mathrm{S}}}\sum_{j=1}^{m_{\mathrm{T}}}\hat{w}(\bx_i)\:\langle\Phi_{\bx_i},\Phi_{\bx_j'}\rangle
	+\frac{1}{m_{\mathrm{T}}^2}\sum_{i,j=1}^{m_{\mathrm{T}}}\langle\Phi_{\bx'_i},\Phi_{\bx'_j}\rangle.
	\end{split}
\end{equation}
We can drop the last term because it is independant of $\hat{w}$. We are thus left with the following quadratic expression in $\hat{w}:=\left(\hat{w}(\bx_1),...,\hat{w}(\bx_{m_{\mathrm{S}}})\right)^\mathsf{T}$:
\begin{equation}
	\label{MMD_quadr}
	\left(\mathrm{MMD}[\hat{p}_{\mathrm{T}},\hat{w}\hat{p}_{\mathrm{S}}]\right)^2 =	
	\frac{1}{m_{\mathrm{S}}^2}\left(\frac{1}{2} \hat{w}^\mathsf{T}K\hat{w} - k^\mathsf{T}\hat{w}\right)+\mathrm{const}
\end{equation}
where $K_{ij}:=2K(\bx_i,\bx_j)$ and $k_i:=\frac{2m_{\mathrm{S}}}{m_{\mathrm{T}}
}\sum_{j=1}^{m_{\mathrm{T}}}K(\bx_i,\bx_j')$. Experience has shown \cite{Cov_shift_by_KMM} that imposing the following two constraints on the weights $\hat{w}$ in the minimizing problem helps in finding a good solution:
\begin{equation}
	\label{quadratic_problem}
	\min_{\hat{w}}\left(\frac{1}{2} \hat{w}^\mathsf{T}K\hat{w} - k^\mathsf{T}\hat{w}\right)
	\textnormal{  with  }
	\hat{w}(\bx_i) \in [0,B] 
	\textnormal{  and  } 
	\left|\frac{1}{m_{\mathrm{S}}}\sum_{i=1}^{m_{\mathrm{S}}}\hat{w}(\bx_i)-1\right|\leq \epsilon
\end{equation}
where $\epsilon=O(B/\sqrt{m_{\mathrm{S}}})$. The condition $\hat{w}(\bx_i) \in [0,B]$ ensures that $\hat{p}_{\mathrm{S}}$ and $\hat{p}_{\mathrm{T}}$ are not too far apart and limits the influence of single observations. The $|...|<\epsilon$ condition ensures that the corrected distribution $\hat{w}\hat{p}_{\mathrm{S}}$ is indeed close to a probability distribution. 

From here on, the only thing we need is a good quadratic program to solve (\ref{quadratic_problem}) under the numerical conditions at hand.
	
\subsubsection{In practice}
\label{cov_shift_in_practice}
Numerous Python API's are available to solve the quadratic problem (\ref{quadratic_problem}). The \emph{Quadratic Programming in Python} web page\footnote{\texttt{https://scaron.info/blog/quadratic-programming-in-python.html}} proposes a list of QP solvers and gives a comparative study of their performance.
\\

The Breast Cancer dataset\footnote{\texttt{https://archive.ics.uci.edu/ml/datasets/Breast+Cancer+Wisconsin+(Diagnostic)}} was used in \cite{Cov_shift_by_KMM} to compare the efficiency of the KMM method with an unweighted baseline method on a simple classification task. The data set contains 699 descriptions of tumors, each characterized by 10 integer valued features $\bx$ and classified as either benign $y=1$ or as malignant $y=0$. This is a non linear classification problem for which a SVM with a Gaussian kernel $K_{ij}=\exp[-\Vert\bx_i-\bx_j\Vert^2/(2\sigma_{\mathrm{SVM}}^2)]$ turns out to be well suited \cite{Bishop}. A range of values $C\in\{0.01,0.1,1,10,100\}$ was tested for hyperparameter which defines the smoothness of the boundary of the classifier. Smaller values mean a stronger regularization and thus a smoother boundary. The kernel size was fixed to $\sigma^2_{\mathrm{SVM}} =5$. 
\\

\begin{figure}[h]
\centering
\includegraphics[scale=0.28]{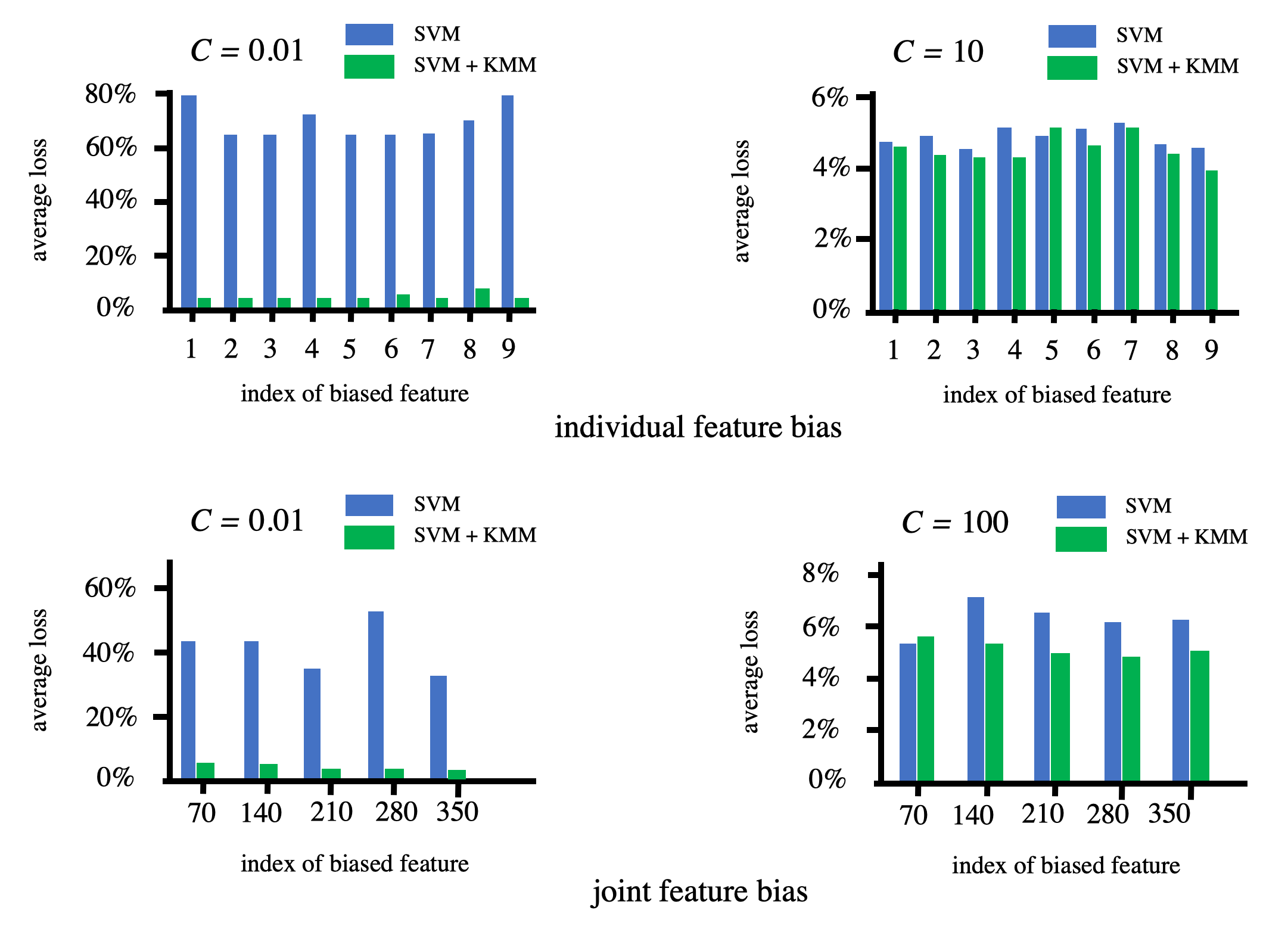}
\caption{{\small The top row compares the prediction errors of an ordinary SVM with those of a KMM-corrected SVM in the case of an individual feature bias. The lower row shows the corresponding prediction errors for a joint feature bias.}}
\label{results_KMM_fig} 
\end{figure}

Two types of sample selection biases where introduced after performing an initial split of the available data into a train set and a test set: 
\begin{enumerate}
\item In the first case, an individual feature bias was intentionnaly introduced in the test set by subsampling observations with probability 0.2 when $x_j\leq 5$ and with probability 0.8 when $x_j> 5$, separately for each of the 10 features.
\item In the second case, a joint feature bias was introduced by keeping an observation $(\bx,y)$ in the training set with a probability $\propto\exp(-\gamma\Vert \bx-\bar{\bx}\Vert^2)$ with $\gamma=1/20$ which decreases as points are further away from the sample mean $\bar{\bx}$. 
\end{enumerate}
Solving for the MMD reweighting $\hat{w}$ with (\ref{quadratic_problem}) introduces three more hyperparamters $\varepsilon, B$ and $\sigma$. The following choices were made: $\varepsilon=(\sqrt{m_{\mathrm{S}}}-1)/\sqrt{m_{\mathrm{S}}}$, $B=1000$ was chosen such that all $\hat{w}(\bx_i)$ are below this limit and $\sigma$ was chosen to equal the SVM kernel size $\sigma_{\mathrm{SVM}}$ with no better reason than intuition. Results are shown in figure \ref{results_KMM_fig}.
A number of comments can be made. The KMM method looks particularly efficient for reducing the error in the presence of a covariate shift when $C$ is small ($\approx 0.01$), that is for “simple” models. It also reduces the error moderately when $C$ is very high ($\approx 10-100$). For intermediate values of $C=O(1)$ however it can make things slightly worse. Finding a principled way to chose an appropriate KMM-kernel remains an open question. The same is true for the problem of determining which properties are favorable for the application of the KMM method. 
\\

As a conclusion we can say that the KMM method for dealing with a covariate shift seems to works well for simpler models. However it is certainly not a one-size-fits-all solution that we could apply blindly. It is an option to try and whose efficiency has to be verified. 

\subsection{Concept shift}
\label{concept_shift_subsection}

\subsubsection{Intuition}
\begin{figure}[h]
\centering
\includegraphics[scale=0.3]{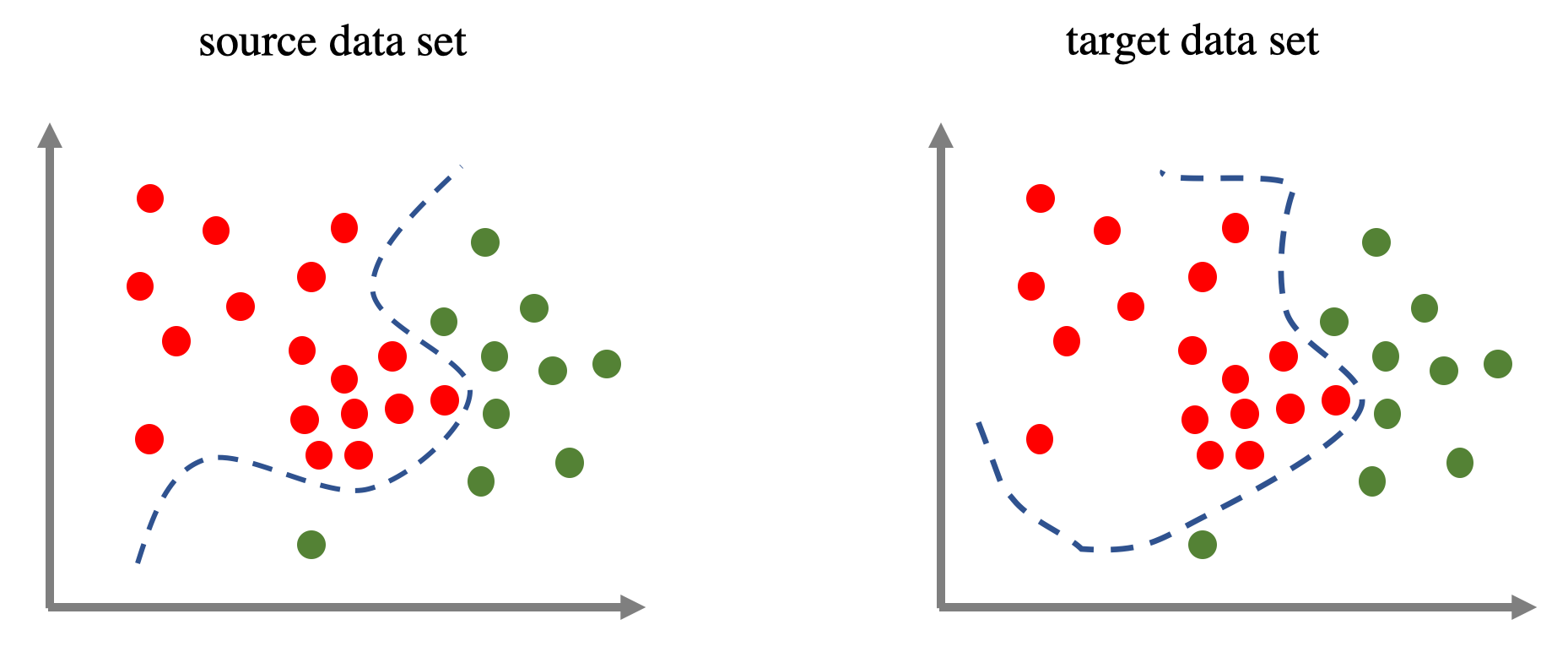}
\caption{{\small In a concept shift the conditional distributions $p_{\mathrm{S,T}}(y|\bx)$, depicted as a dashed frontier, are different in the source and in the target while the prior distributions $p_{\mathrm{S,T}}(\bx)$ are equal.}}
\label{concept_shift_fig} 
\end{figure}

Concept shift, or real concept shift as it is sometimes referred to \cite{A_survey_on_CDA}, defines a situation in which the dependence of the target variable $y$ on the features $\bx$ is different in the source and in the target. In other words the conditional distributions $p_{\mathrm{S}}(y|\bx)\neq p_{\mathrm{T}}(y|\bx)$ differ. The prior distributions over features are however supposed to coincide, $p_{\mathrm{S}}(\bx)=p_{\mathrm{T}}(\bx)$. Figure \ref{concept_shift_fig} depicts the situation symbolically.
\\ 

\begin{figure}[h]
\centering
\includegraphics[scale=0.25]{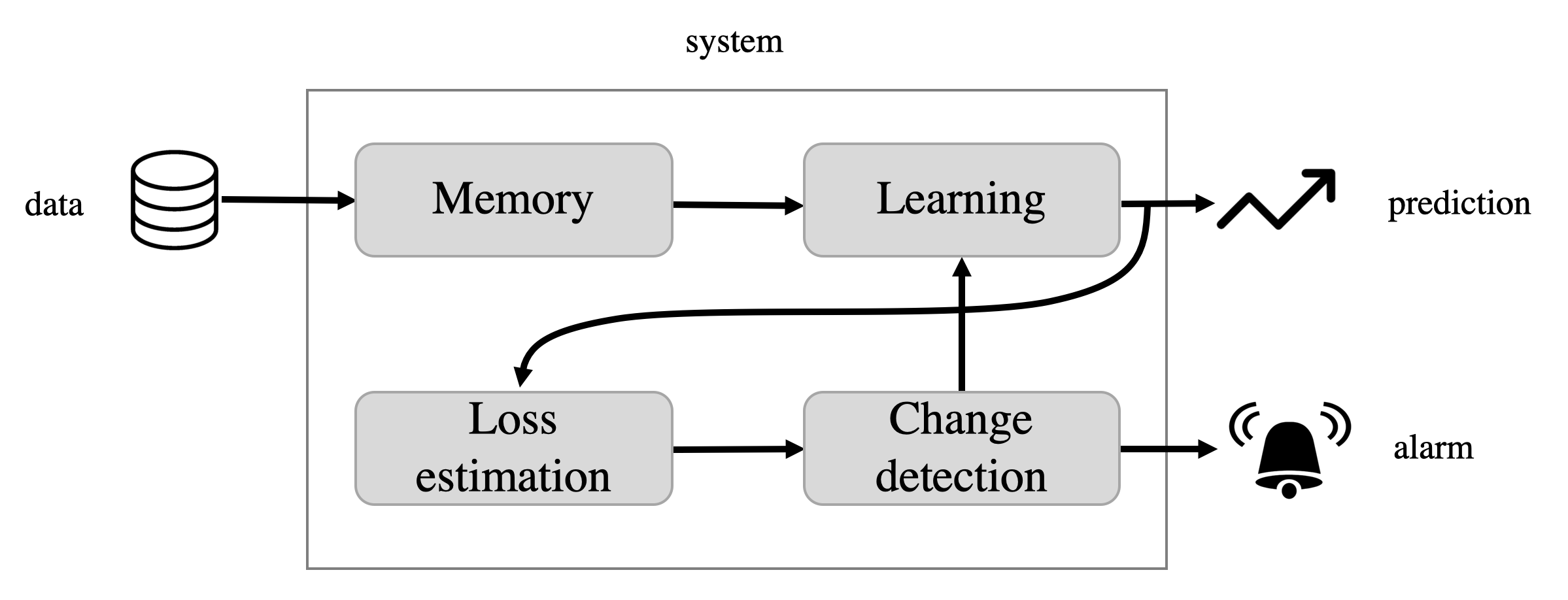}
\caption{{\small The four modules of an adaptive system.}}
\label{concept_shift_modules_fig} 
\end{figure}

Just like for the prior shift that we discussed in section \ref{prior_shift_section}, we generally know nothing about the target labels here either. Therefore we cannot simply deal with a concept shift by defining a reweighting factor $w:=\frac{p_{\mathrm{T}}(y|\bx)}{p_{\mathrm{S}}(y|\bx)}$ in the definition (\ref{true_Risk}) of the true risk that we want to minimize. Concept shift often takes the form of a \emph{concept drift}, where the conditional probability $p_{t}(y|\bx)$ depends explicitly on the time $t$, in other words it occurs in \emph{non stationary} environments, for instance in aging systems whose dynamics changes progressively. In other cases however the change could be sudden, recurrent or even a mixture of these cases. Learning in the presence of a concept drift can also be regarded as a generalization of \emph{incremental learning}  where a system learns one example at a time. Algorithms that deal with drift are globally termed \emph{adaptive algorithms} as they are designed to dynamically adapt to evolving settings. The core of the difficulty one faces in such dynamic environments is to be able to distinguish a random outlier of the response variable $y$ from a genuine drift in the relationship which binds $\bx$ and $y$. Here are a few basic strategies that can be used to deal with a concept shift:
\begin{enumerate}
\item Periodically retrain the model with fresh data selected from a sliding time window.
\item Periodically update the model with fresh data while keeping part of the old data.
\item Weight training data inversely proportional to their age, if the algorithm allows to do so.
\item Use an iterative scheme were a new model learns to correct the most recent model (similar to a boosting).
\item Detect drift and select a new model accordingly.
\end{enumerate}
Organizing the wealth of existing methods that deal with the concept shift problem a challenge by itself. One way to do this has been proposed in \cite{A_survey_on_CDA}. It considers that any such adaptive system has four parts as figure \ref{concept_shift_modules_fig} shows:
\begin{itemize}
\item the \emph{Learning Algorithm Module} that actually gets trained,
\item a \emph{Memory Module} that defines which data is presented to the learning algorithm,
\item a \emph{Loss Estimation Module} tracks the performance of the learning algorithm,
\item the \emph{Change Detection Module} processes the information provided to it by the loss estimation module and updates the Learning Algorithm when necessary.
\end{itemize}
An in-depth description of all options for these four modules is outside the scope of this introduction. We therefore refer the interested reader to \cite{A_survey_on_CDA} for a thorough discussion.

\subsubsection{What mathematics tells us}
As previous section shows, a concept shift can take many different forms in practice. For this reason mathematics unfortunately does not have much to tell in general. One exception though is the situation where the drift is continuous and slow enough for a model to progressively adapt to the change. This setting can indeed be formalized within an extension of the PAC theory (sketched in section \ref{PAC_learning_subsection}) that we quickly present here.
\\

Recall that the basic PAC theory defines what it means for a hypothesis class $\mathcal{H}$ to be \emph{learnable}. For this we briefly refer for to our discussion below equation (\ref{true_Risk}):
\begin{quote}
“There should exist an algorithm $A$ which takes observations $S$ sampled from the distribution $p$ as input and selects with a high confidence a predictor $\hat{h}_S=A(S)$ whose true risk $R[\hat{h}_S]$ is close to the best that can be achieved within a given hypothesis class, provided the sample size $m=|S|$ is large enough.”
\end{quote}
On intuitive ground we expect that a similar result should hold when  the drift is slow enough. This can indeed be proven within a formalization proposed in \cite{Trackin_DC}. It considers a binary classification problem with a continuous drift. It first assumes that the true dependence of the target $y$ on the features $\bx$ is given by a deterministic function $h_t:\mathcal{X}\rightarrow\mathcal{Y}=\{0,1\}$ which evolves over time $t$. The second assumption is that this target moves slowly, in other words that $h_t$ should somehow be close to $h_{t-1}$. This is formalized as the requirement that the probability of two successive predictors $h_t$ and $h_{t+1}$ making different predictions for features $\bx$ sampled according to $p$ is small: 
\[
	p[h_{t+1}(\bx)\neq h_t(\bx)] \leq\Delta.
\]
A \emph{tracking strategy} is then defined as an algorithm $A$ which takes the last $m$ observations $S_m:=\left((\bx_1,h_1(\bx_1),..., (\bx_m,h_m(\bx_m))\right)\in\left(\mathcal{X}\times\mathcal{Y}\right)^m$ of the system as an input to construct a predictor $\hat{h}_{S_m}:=A(S_m)$ for guessing the next value $h_{m+1}(\bx_{m+1})$ using $\hat{h}_{S_m}(\bx_{m+1})$. Paralleling the original definition (\ref{PAC_true_risk}) of a learnable hypothesis class $\mathcal{H}$, we now define  what it means for $\mathcal{H}$ to be \emph{trackable}. Assuming the sampling distribution for features $\bx$ is $p$, $\mathcal{H}$ is trackable if there exists an algorithm $A$ such that the true risk $R[\hat{h}_{S_m}]:=p[\hat{h}_{S_m}(\bx)\neq h_{m+1}(\bx)]$ is small, when averaged over samples $S_m:=\left((\bx_1,h_1(\bx_1),..., (\bx_m,h_m(\bx_m))\right)$ where each $\bx_i$ is drawn from $p$:
\begin{equation*}
\begin{split}
	\E_{(\bx_1,...,\bx_m)\sim p^{m}}\:\:R\left[\hat{h}_{S_m}\right] &\leq\epsilon 
\end{split}
\end{equation*}
provided that the change rate $\Delta$ is small enough and that the sample size $m$ is large enough. With these definitions in place, the following can be proved \cite{Trackin_DC} under some technical assumptions. There indeed exists a (randomized) tracking strategy $A$ as long as $\Delta<\epsilon$ and:
\begin{align}
	\label{bound_m}
	m &> \mathrm{const}\:(d/\epsilon)\log(1/\epsilon), \\
	\label{bound_eps}
	\Delta &< \frac{\mathrm{const}\:\epsilon^2}{d\log 1/{\epsilon}},
\end{align}
where “$\mathrm{const}$” stands for a numerical constant. Equation (\ref{bound_m}) confirms that the sample size $m$ grows both with the hypothesis VC-dimendsion $d$ of the hypothesis class and with the accuracy requirement $1/\epsilon$ on the predictions. Equation (\ref{bound_eps}) on the other hand shows that the higher the complexity $d$ and the accuracy $1/\epsilon$, the slower the drift $\Delta$ should be.

\subsubsection{In practice}
The number of applications that deal with concept drift is so large that whole studies have been devoted solely to categorize such tasks into a coherent framework. The survey \cite{An_overview_of_CDA} for instance distinguishes three groups of applications, that have different goals and use different kind of data:
\begin{enumerate}
\item \textbf{Monitoring and control applications}. Applications in this group aim at monitoring some automated production process in order to measure its quality or to detect possible outliers. Changes  occur fast, typically within minutes or seconds.
\item \textbf{Information management applications}. Applications in this group aim at organizing information. Typically data comes in as a sequence of time stamped web documents that should be classified or characterized. Drift could happen gradually of suddenly over several days or weeks. 
\item \textbf{Diagnostics applications}. Tasks in this group aim at characterizing the health of populations of economies that happen over months or years and are gradual.
\end{enumerate}
Each of these applications in each category generally use different techniques to deal with the concept shift they face. As it would not make much sense in this review to single out one of these as being particularly representative, we refer the interested reader to the literature for how concept shift is dealt with in specific cases \cite{An_overview_of_CDA,  Understand_CD, Survey_on_MDD}.

\subsection{Subspace mapping}
\label{domain_shift_subsection}

\subsubsection{Intuition}
\begin{figure}[h]
\centering
\includegraphics[scale=0.25]{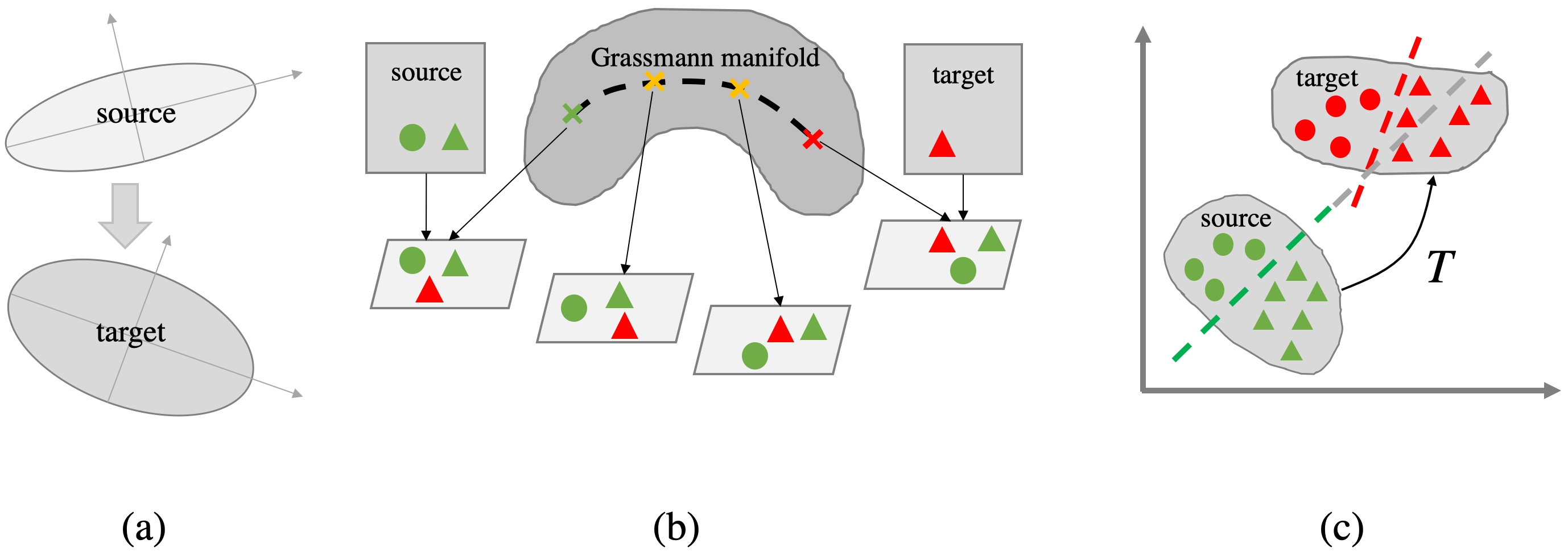}
\caption{{\small (a) A elementary method looks for an affine transformation between source and target subspaces defined by their respective PCA, (b) the methods of geodesic on Grassmann manifolds looks for a sequence of transformations between the subspaces defined by the PCA's on the source and target domains}, (c) optimal transport methods looks for a “low cost” non-linear transformation $T$ from the source to the target.}
\label{subspace_mapping_3cases} 
\end{figure}

Let's not deviate from tradition and imagine  we want to train a classifier to recognize cats and dogs. Assume that our training set contains pictures shot under specific lighting conditions which include, say, exposure, color balance and a specific kind of background. How then should we optimize our classifier to classify images of these same pets shoot under very different lighting conditions? This is typically a problem of \emph{subspace mapping} \cite{intro_to_DA}. More generally, subspace mapping deals with situations where the source and target examples are selected likewise but there is an unknown twist, or coordinate change $T$, between the features $\bx$ describing these in the source and those describing them in the target $\bx'=T(\bx)$. More precisely, in terms of the joint probability distributions we assume that for a subspace mapping the following holds:
\begin{equation}
	\label{pS_to_pT_sm}
	p_{\mathrm{T}}(T(\bx),y)=p_{\mathrm{S}}(\bx,y)
\end{equation}
for some unknown distortion $T$ between the descriptions in the source and target domains. For simplicity we shall assume henceforth that $T:\mathcal{X}\rightarrow\mathcal{X}$, but nothing in principle prevents the target feature space to be different from the source feature space. By marginalizing and conditioning (\ref{pS_to_pT_sm}) we respectively obtain:
\begin{align}
	p_{\mathrm{T}}(T(\bx)) &= p_{\mathrm{S}}(\bx), \label{pS_to_pT_sm_f} \\
	p_{\mathrm{T}}(y\vert T(\bx)) &= p_{\mathrm{S}}(y\vert \bx) \label{pS_to_pT_sm_c}. 
\end{align}
In words: predictions $y$ match for observations described by $\bx$ in the source and by $T(\bx)$ in the target. Equation (\ref{pS_to_pT_sm_f}) implies that $p_{\mathrm{T}}(\bx)\neq p_{\mathrm{S}}(\bx)$ and equation (\ref{pS_to_pT_sm_c}) implies $p_{\mathrm{T}}(y\vert \bx) \neq p_{\mathrm{S}}(y\vert \bx)$. Thus, another way to look at subspace mapping is as a composition of a covariate shift with a concept shift both characterized by a change of coordinates $T$ on features. There is a wide range of methods that try to to deal with a subspace mapping. Lets describe three of them quickly in intuitive terms as shown in figure \ref{subspace_mapping_3cases}.
\begin{itemize}

\item \textbf{Principal Component Analysis}: This is the simplest method. It looks for an affine transformation that maps the $n$ principal components of the source data into the $n$ principal components of the target data (with $\mathcal{X}\subset\R^d$ and $n<d$).

\item \textbf{Geodesics on Grassmann Manifolds}: Rather than looking like above for a one-shot transformation, this methods looks for a sequence of intermediate $n$ dimensional representations that smoothly interpolate between the source and target representations. Mathematically, each representation corresponds to a projection of the source and target data onto an intermediate $n$ dimensional subspace of $\R^d$ regarded as a point along a geodesic in an $n\times d$-dimensional space know as a Grassmann manifold. This method is inspired by incremental learning.

\item \textbf{Optimal transport (OT)}: This is currently the most generic, elegant and powerful method for dealing with subspace mapping. In short, optimal transport looks for a mapping that morphs the source sample distribution $\hat{p}_{\mathrm{S}}$ into the target distribution $\hat{p}_{\mathrm{T}}$ while using minimal effort to move mass from one to the other. Although the mathematics of OT has already been know for decades, it has long been ignored by data scientists, mainly due to its heavy computational cost. But things have changed lately due to various computational tricks that allow for fast optimization. Sounds interesting? So, take a deep breath and jump to next section!
\end{itemize}

\subsubsection{What mathematics tells us on optimal transport}
Optimal transport is a beautiful mathematical theory that has a long history going back to the $18^{\rm{th}}$ century with seminal work by G. Monge and, more recently, in the 1940's by L. Kantorovich. For the curious reader we refer for instance to the excellent book \cite{Comp_opt_trans}. Our purpose here is to introduce the strictly minimal number of concepts that will allow us to apply OT to subspace mapping. In a nutshell, optimal transport defines a discrepancy measure between two probability distributions $\alpha$ and $\beta$ as the minimal effort required to move mass, bit by bit, from one to the other, hence its name: the \emph{earth-mover's distance}. This is to be contrasted with the classical measures like the KL divergence or the MMD distance we mentioned earlier. These compare two distributions, either by comparing their values $\alpha(\bz)$ and $\beta(\bz)$ at the same point $\bz$ (for KL for instance), or by comparing them at nearby points (as defined by a kernel $K$ for the MMD distance in (\ref{MMD_empirical}) for instance). 
\\

Still another way to look at OT is to regard it as a way to generalize a notion of distance $d$ between points $\bz$ and $\bz'$ in a space $\Omega$ towards a distance measure between probability distributions $\alpha$ and $\beta$ on that space. This new distance, known as the \emph{Wasserstein distance} $W_p(\alpha,\beta)$ between $\alpha$ and $\beta$, is a \textit{lifting} of $d$ from $\Omega$ onto the space of distributions on $\Omega$. Indeed, for point measures\footnote{Depending on the context, these point measure should be understood either as Kronecker deltas or as Dirac measures.} $\alpha=\delta_\bz$ and $\beta=\delta_{\bz'}$ the distance $W_p(\delta_\bz, \delta_{\bz'})$ reduces to $d(\bz, \bz')$  as we shall see.
\\

For the sake of simplicity we restrict ourselves to discrete measures on a space $\Omega\subset\R^n$, but everything we say can be generalized to arbitrary measures \cite{Comp_opt_trans}. Later on $\Omega$ will denote either our feature space $\mathcal{X}$ or the full observation space $\mathcal{X}\times\mathcal{Y}$. 

\begin{figure}[h]
\centering
\includegraphics[scale=0.15]{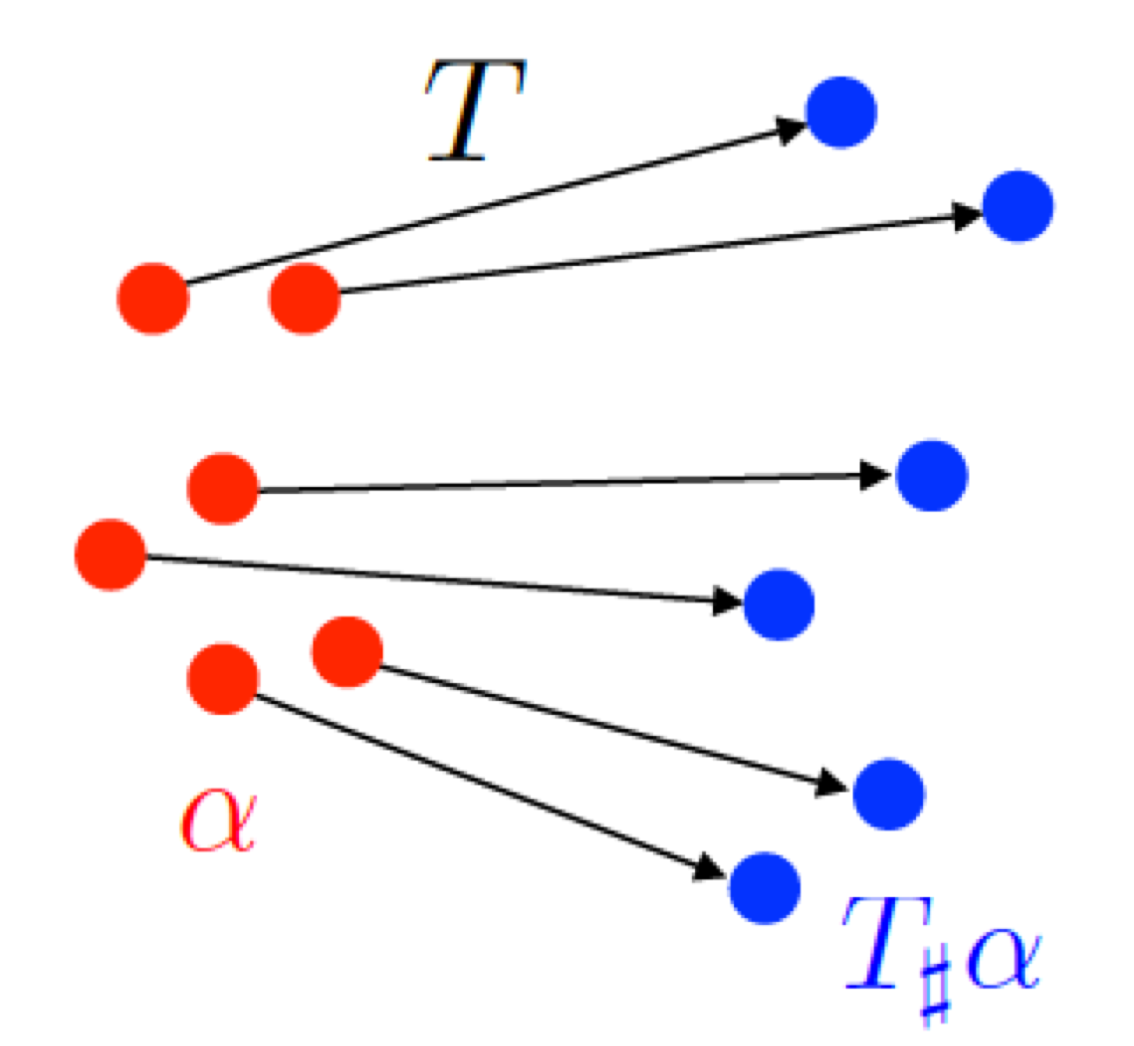}
\caption{{\small The push-forward mapping $T_\#$ acting on a discrete distribution just moves the support points by $T$.}}
\label{push_forward} 
\end{figure}

\paragraph*{The Wasserstein distance}
The first step is to define how a measure $\alpha$ can be transported by a mapping $T:\Omega\rightarrow\Omega$. Let 
\[
	\alpha :=\sum_{i=1}^m \alpha_i \,\delta_{\bz_i}
\]
be a measure whose support is defined by a set of discrete points $\bz_i$ and with weights $\alpha_i>0$ that sum to $1$. We define a transport (or push-forward) operator $T_\#$ on distributions by simply shifting the support points of $\alpha$ by $T$:
\[
	T_\# \alpha :=\sum_{i=1}^m \alpha_i \,\delta_{T(\bz_i)},
\]
This is illustrated in figure \ref{push_forward}.

\begin{figure}[h]
\centering
\includegraphics[scale=0.15]{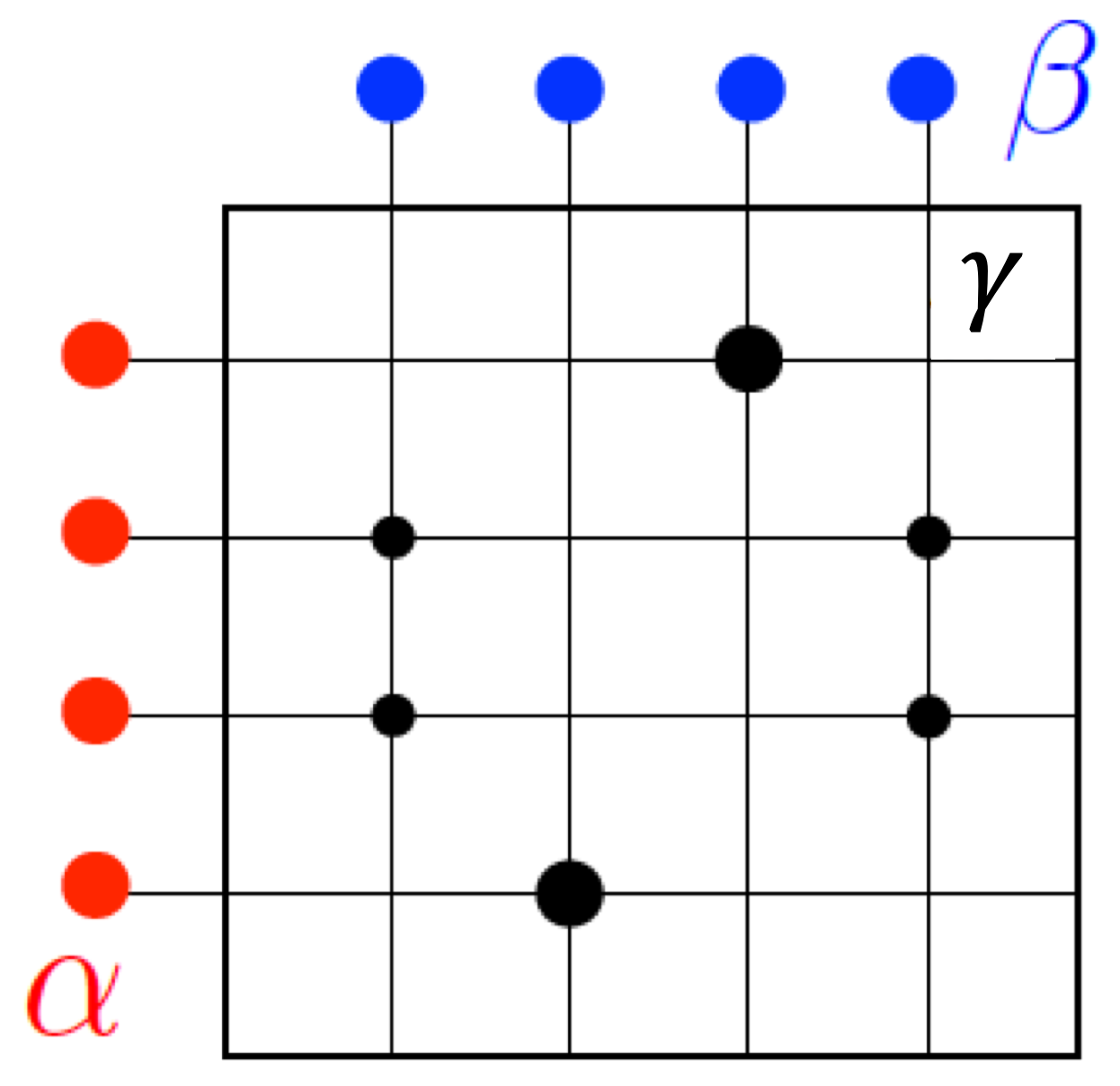}
\caption{\small The Kantorovich formulation of OT assume the mass at each point in the support of $\alpha$ is redistributed onto the whole support of $\beta$.}
\label{OT_transport_plan_1D} 
\end{figure}

Now assume we are given a cost function $c$ that measures the effort required to move a unit of mass from a location $\bz$ to another location $\bz'$. The corresponding effort to move the whole mass from $\alpha$, bit by bit, to $T_\#\alpha$ is then naturally defined by
\begin{equation}
	\label{push_forward_cost}
	\E_{\bz\sim\alpha}[c(\bz,T(\bz))]:=\sum_{i=1}^m \alpha_i\,c(\bz_i, T(\bz_i)).
\end{equation}
Now, assume the target distribution $\beta$ is given and consider all maps $T$ which move $\alpha$ into $\beta$, that is  all maps such that $\beta =T_\#\alpha$. The most economic such $T$ is thus defined as
\begin{equation}
	\label{min_push_forward_cost}
	\argmin_{T: T_\#\alpha = \beta} \E_{\bz\sim\alpha}[c(\bz,T(\bz))].
\end{equation}
Finding such a minimizer $T$ is known as the \emph{Monge problem}. In its full generality this is an exceedingly hard problem, especially because the constraint on $T$ is non-convex. For this reason, following a suggestion by Kantorowitch, the above problem is generally relaxed. Rather than looking for a mapping $T$ that deterministically maps a point $\bz$ in the support of $\alpha$ to a point $T(\bz)$ in the support of $\beta$, we look for a so-called probabilistic \emph{transport plan} $\gamma$ that redistributes the mass $\alpha_i$ at a point $\bz_i$ in $\alpha$ onto the whole support of $\beta$ as figure \ref{OT_transport_plan_1D} shows. 
Formally, if $\lbrace \bz_i \rbrace_{i=1}^r$ and $\lbrace \bz'_j \rbrace_{j=1}^{s}$ are the respective supports of $\alpha$ and $\beta$, the transport plan $\gamma$ is defined as a probability distribution on $\Omega\times\Omega$:
\begin{equation}
	\label{gamma_trans_plan}
	\gamma := \sum_{i,j}\gamma_{ij}\,\delta_{(\bz_i, \bz'_j)}
\end{equation}
satisfying the two following marginality constraints:
\begin{align}
	\label{marg_alpha}
	\sum_{j=1}^s \gamma_{ij} &= \alpha_i, \\
	\label{marg_beta}
	\sum_{i=1}^r \gamma_{ij} &= \beta_j.
\end{align}
Paralleling (\ref{push_forward_cost}), the cost of the transport plan $\gamma$ under a unit cost $c$ is now defined as:
\begin{equation}
	\label{plan_cost}
	\E_{(\bz,\bz')\sim\gamma}[c(\bz,\bz')]:=\sum_{i,j} \gamma_{ij}\,c(\bz_i, \bz_j')
\end{equation}
Let $U(\alpha,\beta)$ denote the set of transport plans $\gamma$ that satisfy the constraints (\ref{marg_alpha}) and (\ref{marg_beta}). The minimal transport cost for moving $\alpha$ towards $\beta$ is thus defined as:
\begin{equation}
	\label{min_plan_cost}
	{\cal{L}}_c(\alpha,\beta):=\min_{\gamma:\gamma\in U(\alpha,\beta)} \E_{(\bz,\bz')\sim\gamma}[c(\bz,\bz')].
\end{equation}
Equations (\ref{plan_cost}) and (\ref{min_plan_cost}) are known as the \emph{Kantorovich problem}\footnote{When the support of the optimal plan $\gamma$ is concentrated on a one-dimensional path $\lbrace(\bz_i,\bz_i'=T(\bz_i))_{i=1}^m\rbrace\subset\Omega\times\Omega$ then the Kantorovich solution is also a solution to the Monge problem, but this need not always be the case.}. Now comes the crucial point. If $d$ is a metric on $\Omega$ and if we define a cost function by $c(\bz,\bz'):=\left[d(\bz,\bz')\right]^p$ for some $p\geq 1$, then the Wasserstein distance between $\alpha$ and $\beta$ is defined as:
\begin{equation}
	\label{Wassertein_def}
	W_p[\alpha,\beta] := \left[{\cal L}_{d^p}(\alpha,\beta)\right]^{1/p}.
\end{equation}
This quantity indeed deserves to be termed a distance because it can be proved that it is positive, symmetric and satisfies the triangle inequality \cite{Comp_opt_trans}. This is the earth-mover distance we were looking for. Let's now see how we can put it to good use for solving the subspace mapping shift.

\paragraph*{Applying OT to the subspace mapping shift}
Let's recall the situation in the presence of a subspace mapping shift that is depicted in figure \ref{subspace_mapping_3cases}(c):
\begin{itemize}
\item We assume we have a train set $\lbrace(\bx_i,y_i)_{i=1}^{m_{\mathrm{S}}}\rbrace$ sampled from a source distribution $p_{\mathrm{S}}(\bx,y)$. Henceforth this sample will be assimilated with the an empirical distribution $\hat{p}_{\mathrm{S}}(\bx,y)$.
\item We suppose that the target distribution obeys $p_{\mathrm{T}}(T(\bx),y)=p_{\mathrm{S}}(\bx,y)$ for some unknown distortion mapping $T$.
\item We have a set of features $\lbrace(\bx_j')_{j=1}^{m_{\mathrm{T}}}\rbrace$ from observations in the target domain sampled from $p_{\mathrm{T}}(\bx')$ for which we have no labels, however. The corresponding empirical marginal is denoted $\hat{p}_\mathrm{T}(\bx)$.
\end{itemize}
There are various options to apply OT to this shift setting. The main alternative is between applying OT on $\Omega=\mathcal{X}$ to transport the marginal $p_{\mathrm{S}}(\bx)$ towards $\hat{p}_{\mathrm{T}}(\bx')$ or, alternatively, to apply OT on $\Omega=\mathcal{X}\times\mathcal{Y}$ to transport the joint distribution $\hat{p}_{\mathrm{S}}(\bx,y)$ towards the, so far unknown, joint distribution $\hat{p}_{\mathrm{T}}(\bx',y')$.  
\\

The first option is probably the most straightforward at first sight. Given some metric $d$ on our feature space $\cal{X}$ and some $p\geq 1$ we would look for an OT plan $\gamma$ that transforms $\hat{p}_{\mathrm{S}}(\bx)$ into $\hat{p}_{\mathrm{T}}(\bx')$. To preserve label information during transport, we should also somehow penalize plans $\gamma$ which move source observations $(\bx_1,y_1)$ and $(\bx_2,y_2)$ with different labels $y_1\neq y_2$ to the same point $\bx'$ in the target. Once we have computed such a transport plan $\gamma$, we could use it to move our labeled source samples into the target domain and train a ML model on those transported versions of the source observations. This strategy was explored indepth in \cite{OT_for_DA}, it works, but the penalization procedure remains somehow awkward and ad hoc.
\\

The second option is the one we are going to describe in more detail below. It removes some of the previous arbitrariness and its theoretical justification is more sound. It is also more efficient in practice and was pioneered in \cite{Joint_distrib_OT}. The basic idea is rather natural. First, we need a distance $d$ between observations in $\mathcal{X}\times\mathcal{Y}$ to define $W_p$. So let's assume we have one, say $\rho(\bx,\bx')$, on $\mathcal{X}$ and that we also have a loss function $\ell(y,y')$ on $\mathcal{Y}$ to compare labels. So we can for instance define:
\begin{equation}
	\label{def_base_distance}
	d\left((\bx,y),(\bx',y')\right):=\lambda\,\rho(\bx,\bx')+\ell(y,y'),
\end{equation}
where $\lambda$ balances the ratio between the distance $\rho$ and the loss $\ell$. Now that we are equipped with this distance $d$ it is tempting to use it to find the corresponding optimal transport plan $\gamma$ that maps $\hat{p}_{\mathrm{S}}(\bx,y)$ towards $\hat{p}_{\mathrm{T}}(\bx',y')$ in order to correct the training process of our ML model. But wait, in the target domain we only have a sample $\lbrace(\bx_j')_{j=1}^{m_{\mathrm{T}}}\rbrace$ of \emph{unlabeled} features ! How on earth could we approximate $\hat{p}_{\mathrm{T}}(\bx',y')$? Well, remember that, in the end, the only thing we really care about is finding a good predictor $h$ for the labels $y'=h(\bx')$ in the target domain. If we had such a predictor $h$, we could use it to approximate the target distribution with
\[
	\hat{p}_{\mathrm{T}}^h(\bx',y') := \hat{p}_{\mathrm{T}}(\bx')\,\delta_{y'=h(\bx')},
\]
where $\hat{p}_{\mathrm{T}}(\bx')$ is the empirical marginal distribution that we do know. In words, we could approximate the true dependency of $y'$ on $\bx'$ in the target with the predictions $y'\approx h(\bx')$ made using $h$. Now, to find a good predictor we can look for the one which will minimizes the Wasserstein distance $W_1$ between the \emph{joint} source distribution $\hat{p}_{\mathrm{S}}(\bx,y)$ and the approximate target distribution $\hat{p}_{\mathrm{T}}^h(\bx',y')$ defined with $h$:
\begin{equation}
	\label{h_OT}
	h_{\mathrm{OT}} := \argmin_{h\in\mathcal{H}} W_1(\hat{p}_{\mathrm{S}},\hat{p}^h_{\mathrm{T}}).
\end{equation}
The value $p=1$ in (\ref{h_OT}) is chosen on a theoretical ground. As a matter of fact, it can be shown that it is $W_1$ which specifically occurs in a generalized PAC bounds on the true expected risk \cite{Joint_distrib_OT}. Solving (\ref{h_OT}) involves a double minimizing process, one for the transport plan $\gamma$ and another for the predictor $h$. Using (\ref{plan_cost}), (\ref{min_plan_cost}) and (\ref{Wassertein_def}) with $p=1$ and $c=d^1$, the optimal $h$ in (\ref{h_OT}) is defined by:
\begin{equation}
	\label{double_min}
	h_{\mathrm{OT}}=\argmin_{\substack{h\in		
	\mathcal{H} \\ \gamma\in U(\hat{p}_{\mathrm{S}},\hat{p}_{\mathrm{T}}^h)}}
	\sum_{i,j} \gamma_{ij}\,d((\bx_i, y_i), (\bx_j',h(\bx_j')))
\end{equation}
As is usual in ML, a regularization term $\varepsilon\,\mathrm{reg}[h]$ is added to prevent $h$ from overfitting the training sample. Expanding $d$ in (\ref{double_min}) according to (\ref{def_base_distance}), we must thus solve the following double minimization problem:
\begin{equation}
	\label{double_min_explicit}
	h_{\mathrm{OT}}=\argmin_{\substack{h\in		
	\mathcal{H} \\ \gamma\in U(\hat{p}_{\mathrm{S}},\hat{p}_{\mathrm{T}}^h)}}
	\left(
		\sum_{i,j} \gamma_{ij}
		\left[\lambda\rho(\bx_i, \bx_j')+\ell(y_i,h(\bx_j'))\right]
		+ \varepsilon\,\mathrm{reg}[h]
	\right)
\end{equation}
To be able to learn $h$, the cost function $\ell$ should obviously be differentiable with respect to its second argument. Both $\lambda$ and $\varepsilon$ are hyperparameters that should be adjusted by a cross-validation. When $\ell =O(1)$, setting $\lambda$ to the inverse of the maximal $\rho$-distance between points sometimes turns out to be a good choice.

\paragraph*{Computing the optimal transport }
Minimizing (\ref{double_min_explicit}) simultaneously with respect to $\gamma$ and $h$ can be done using \emph{Block Coordinate Descent} (BCD). This algorithm basically works by alternatively fixing the predictor $h$ and  optimizing for the transport plan $\gamma$ in (\ref{double_min_explicit}) and then doing the opposite in the next step. Minimizing $\gamma$ in (\ref{double_min_explicit}) for a fixed $h$ can be done using any classical OT solver (see section \ref{in_practive_subsection}). When $\gamma$ is fixed and we minimize (\ref{double_min_explicit}) with respect to $h$ we get the following optimization problem:
\begin{equation}
	\label{minimizing_f_gamma_fixed}
	\argmin_{h\in\mathcal{H}}
	\left(
		\sum_{i,j} \gamma_{ij}\,\ell(y_i,h(\bx_j')) + \varepsilon\,\mathrm{reg}[h]\right)
\end{equation}
which is a variant of the classical learning problem in ML because the true label $y_i$ is compared with the predicted labels $h(\bx_j')$ for \emph{all} points in the target sample but with a loss $\ell$ weighted according to the transport probabilities $\gamma_{ij}$. At first sight this is bad news because the complexity of such a task is $O(m_\mathrm{R}m_\mathrm{T})$ which is quadratic in the sample size rather than linear as in ordinary ML. Fortunately, in the common case of a linear regression where $h(\bx')=\mathbf{w}\cdot\bx'$ with a Ridge regularization, that is when $\ell(y,y')=|y-y'|^2$ and $\mathrm{reg}[h]=\Vert \mathbf{w}\Vert^2$, the optimization (\ref{minimizing_f_gamma_fixed}) reduces to:
\begin{equation}
	\label{Modified_Ridge}
	\argmin_{h\in\mathcal{H}}\frac{1}{m_\mathrm{T}}
	\sum_{j=1}^{m_\mathrm{T}}\vert \hat{y}'_j-h(\bx_j')\vert^2 + \varepsilon\,\Vert\mathbf{w}\Vert^2
	\textnormal{  where  }
	\hat{y}_j':=m_\mathrm{T}\sum_{i=1}^{m_\mathrm{S}}\gamma_{ij}\,y_i
\end{equation}
which is now linear in the sample sizes ($O(m_\mathrm{R}+m_\mathrm{T})$). Equation (\ref{Modified_Ridge}) is an ordinary Ridge regression where the fictive target labels $\hat{y}_j'$ have to be computed using the source labels $y_i$ and the current transport plan $\gamma$. When the BCD algorithm produces a sequence of pairs $\left(\gamma^{(k)},h^{(k)}\right)$ which converges to a limit, it can be shown that this limit is a critical point of the optimization problem (\ref{double_min_explicit}) \cite{Joint_distrib_OT}.

\subsubsection{In practice}
\label{in_practive_subsection}
We will describe an example of a regression task that undergoes a subspace mapping where we can use OT profitably to improve the prediction accuracy. To apply OT in practice for a regression problem we need two kind of tools. One is needed to solve the regularized quadratic regression problem (\ref{Modified_Ridge}) for the optimal $h$ for a given plan $\gamma$. For this, the Ridge API\footnote{\texttt{scikit-learn.org/stable/modules/generated/sklearn.linear\_model.Ridge.html}} from scikit-learn perfectly does the job. Another tool we need is an OT solver to optimize the plan $\gamma$ in (\ref{double_min_explicit}) for a fixed $h$. A good shopping place for this is the Python Optimal Transport (POT) API\footnote{\texttt{pot.readthedocs.io/en/stable/quickstart.html}}. There are methods to compute the Wasserstein distance $W$ between to empirical probability distributions $\alpha$ and $\beta$ (termed “histograms”) and there are also methods to compute the actual optimal plan $\gamma$. One interesting feature offered by the POT API is the availability of an entropic regularized OT solver. This kind of solver has recently allowed to scale OT techniques to much larger data sets than was possible only a few years a go \cite{Comp_opt_trans}. This comes with the slight cost of an additional regularization hyperparameter.
\\

To illustrate the practical usefulness of OT we will describe an indoor localization system based on WiFi signals studied in \cite{Joint_distrib_OT}\footnote{The specific data analyzed in \cite{Joint_distrib_OT} was not made public but similar data can be downloaded from Kaggle: \texttt{https://www.kaggle.com/giantuji/UjiIndoorLoc}}. The aim is to localize a device along a hallway by learning a mapping from the vector $\bx=\{x_1,...,x_d\}$ of signal strengths that the device receives from $d$ fixed access points to its location which was discretized into 119 positions. As it turns out, different devices have different sensitivities which translate into different mappings between the vectors of signal strengths $\bx$ and $\bx'$ they each receive. We are therefore in a typical setting where $OT$ can be used, namely to predict the position of a device B using labeled data from another device A.
\\

To apply OT we must define a distance $d$ between observations as in (\ref{def_base_distance}). In this case $\rho$ is simply the euclidean distance $\Vert \bx-\bx'\Vert$ between signal strength vectors, while the cost $\ell=|y-\hat{y}|^2$ is the squared loss between the true value $y$ of a position and some prediction $\hat{y}$. Therefore the fixed $\gamma$ optimization step reduces to the simple Ridge regression (\ref{Modified_Ridge}). The constant $\lambda$ in (\ref{def_base_distance}) is chosen as the inverse maximal distance between signal strengths vectors $\lambda=1/\max_{i,j}\rho( \bx_i,\bx'_j)$. This makes the contributions of the $\rho$ and $\ell$ terms in $d$ of the same order of magnitude.
\\ 

The results for the accuracy\footnote{A prediction is considered to be accurate in \cite{Joint_distrib_OT} when the predicted position falls within 6 meters of the true position.} ($>98\%$) of the prediction of the position obtained in \cite{Joint_distrib_OT}  outperform all existing methods in this OT problem. This very favorable results for OT can probably be interpreted as the consequence of the fact that OT allows to deal with distribution whose support are very far apart, something that other methods cannot cope with. 
\section{Conclusion}
The need for systematic procedures for dealing with situations where a training data set does not faithfully reflect the real world is still understated by many practitioners despite its practical importance in machine learning. We analyze that this is mainly due to the myriad of methods available and to the lack of a unified and well accepted terminology for describing the different DA scenarios. For these reasons data scientists most often favor various tricks, like guessing an appropriate resampling ratios or generating fictitious data, to build appropriate training sets.
\\
 
The main purpose of this review was to explain that in a number of commonly occurring situations there are indeed efficient methods to perform DA that are based on sound theoretical arguments. We tried to present these in a streamlined but conceptually clear fashion. For a prior shift, the EM algorithm applied to finding prior probabilities in the target domain offers an effective solution within the maximum likelihood approach of ML. The covariate shift on the other hand can efficiently be tackled within the PAC framework. Many reweighing methods are available for this purpose, depending on how we chose to measure the discrepancy between the source and target marginal distributions on features. We explained in some detail how the elegant KMM approach can be applied in this case. For the more general subspace mapping, where the target features have undergone some unknown twist with respect to the source features, the recent progress in computational optimal transport now offer powerful tools. Concept shift and concept drift on the other hand remain mainly the realm of empirical methods that largely depend on the specific applications where they are implemented. Yet, slowly changing targets can be analyzed rigorously within a slightly extended version of the PAC theory.
\\
 
We hope this review can make some contribution in clarifying and spreading awareness of this important topic of data science to a wider audience of data scientists.

\appendix
\section{Appendix: Prior shift with the EM algorithm}
\label{appendix:a}

In this appendix we prove that the iterative procedure (\ref{prior_fixed_point_iter}) for finding the posterior $\hat{p}_\mathrm{T}(\omega|\bx)$ and the prior $\hat{p}_\mathrm{T}(\omega)$ is a straightforward application of the \emph{Expectation Maximization} (EM) algorithm which finds the maximum likelihood solution for problems with latent variables \cite{Bishop}. 
\\

The EM algorithm assumes we are given a parametrized probability distribution $p_{{\boldsymbol{\theta}}}(\bx_1,...,\bx_m;\bz_1,...,\bz_m)$ where the $\bx_i$ are observable features while the $\bz_i$ are unobservable latent variables. EM is an iterative procedure for finding the maximum likelihood solution ${{\boldsymbol{\theta}}}_\mathrm{ML}$ for the corresponding marginal distribution $p_{{\boldsymbol{\theta}}}(\bx_1,...,\bx_m)$ where the latent variables $\bz_i$ have been integrated out. The EM algorithms has 4 steps:
\begin{enumerate}
\item Choose an initial value for ${{\boldsymbol{\theta}}}$ and set ${{\boldsymbol{\theta}}}^{\mathrm{old}}={{\boldsymbol{\theta}}}$.
\item \emph{Expectation step} (E): As the latent $\bz_i$ can not be observed directly, make an educated guess with the available data. In practical terms, use $\bx_1,...,\bx_m$ and the current value of ${{\boldsymbol{\theta}}}^{\mathrm{old}}$ to compute the conditional distribution $p_{\boldsymbol{\theta}^{\mathrm{old}}}(\bz_1,...,\bz_m|\bx_1,...,\bx_m)$.
\item \emph{Maximization step} (M): Update the value of the parameter ${{\boldsymbol{\theta}}}$ using the following definitions:
\[
	\begin{split}
	{{\boldsymbol{\theta}}}^\mathrm{new}&:=\argmax_{\boldsymbol{\theta}}
	Q(\boldsymbol{\theta},\boldsymbol{\theta}^{\mathrm{old}})	\textnormal{ where}\\	
	Q(\boldsymbol{\theta},\boldsymbol{\theta}^{\mathrm{old}}) 
	&:= \sum_{\bz_1,...,\bz_m} p_{\boldsymbol{\theta}^{\mathrm{old}}}(\bz_1,...,\bz_m|\bx_1,...,\bx_m)\log p_{{\boldsymbol{\theta}}}(\bx_1,...,\bx_m;\bz_1,...,\bz_m)
	\end{split}
\]
\item Update the value of ${{\boldsymbol{\theta}}}$ accordingly:
\[
	{{\boldsymbol{\theta}}}^{\mathrm{old}} \rightarrow {{\boldsymbol{\theta}}}^{\mathrm{new}}.
\]
If the convergence\footnote{According to some criteria either on ${{\boldsymbol{\theta}}}$ or on $\log p_{{\boldsymbol{\theta}}}$.} of ${{\boldsymbol{\theta}}}$ has not yet occurred go back to step 2.
\end{enumerate}
The sequence $\boldsymbol{\theta}^{(s)}$ that results from the iterations of these EM steps can be shown to systematically increase the likelihood \cite{Bishop}. We certainly hope it moreover converges to the maximum likelihood parameter $\boldsymbol{\theta}_\mathrm{ML}$.
\\

To apply the EM algorithm to the evaluation of the target priors we make the following identifications. The latent variables $\bz_i:=(z_{i1},...,z_{iK})$ are the one-hot-encodings of the categorical variables $y_i\in\{\omega_1,...,\omega_K\}$. The parameter vector ${{\boldsymbol{\theta}}}$ is the list of unknown prior probabilities in the target: ${{\boldsymbol{\theta}}}:=[p_\mathrm{T}(\omega_1),...,p_\mathrm{T}(\omega_K)]^{\mathsf{T}}$. At last, the full joint distribution is:
\begin{equation}
	\label{joint_pXZ}
	\begin{split}
			p_{{\boldsymbol{\theta}}}(\bx_1,...,\bx_m;\bz_1,...,\bz_m) &:= \prod_{i=1}^m \prod_{k=1}^K \:
	[\hat{p}_\mathrm{T}(\bx_i,\omega_k)]^{z_{ik}}\\
        &= \prod_{i=1}^m \prod_{k=1}^K \:
	[\underbrace{\hat{p}_\mathrm{T}(\bx_i|\omega_k)}_{\displaystyle=\hat{p}_\mathrm{S}(\bx_i|\omega_k)} \,\underbrace{\hat{p}_\mathrm{T}(\omega_k)}_{\displaystyle=\theta_k}]^{z_{ik}},
    \end{split}
\end{equation}
and the log likelihood is:
\begin{equation}
	\label{LL_full}
	\begin{split}
	\log \:&p_{{\boldsymbol{\theta}}}(\bx_1,...,\bx_m;\bz_1,...,\bz_m)=\\
	&\sum_{i=1}^m \sum_{k=1}^K z_{ik} \log \hat{p}_\mathrm{T}(\bx_i|\omega_k)
	+ 
	\sum_{i=1}^m \sum_{k=1}^K z_{ik} \log \hat{p}_\mathrm{T}(\omega_k).
    \end{split}
\end{equation}
Similarly, the conditional probability we need in the (E) step is:
\begin{equation}
	\label{cond_ex_theta}
			p_{{\boldsymbol{\theta}}}(\bz_1,...,\bz_m|\bx_1,...,\bx_m) = \prod_{i=1}^m \prod_{k=1}^K \:
	[\hat{p}_\mathrm{T}(\omega_k|\bx_i)]^{z_{ik}},
\end{equation}
where the $\theta$ dependence of $\hat{p}_\mathrm{T}(\omega_k|\bx_i)$ has been omitted to avoid clutter. The (M) step involves computing the expectation of the log-likelihood (\ref{LL_full}) with respect to the conditional probability (\ref{cond_ex_theta}). Let's write $\E_{\boldsymbol{\theta}}[\:.\:|\bx_1,...,\bx_m]$ for this expectation. Each variable $z_{ik}$ is binary, therefore its expectation simply equals the probability that $z_{ik}=1$. Thus, as the $\bz_i$ are independent, we have: 
\begin{equation}
	\label{exp_z_ik}
	\E_{\boldsymbol{\theta}}[z_{ik}|\bx_1,...,\bx_m]=\hat{p}_\mathrm{T}(\omega_k|\bx_i).
\end{equation}
If we care to make the dependence on $\boldsymbol{\theta}$ explicit in the right member of (\ref{exp_z_ik}) we can use (\ref{prior_fixed_point}) to get:
\begin{equation}
	\label{p_T_omega_xi_iter}
		\hat{p}_\mathrm{T}(\omega_k|\bx_i) = \frac{\displaystyle{\frac{\hat{p}_\mathrm{T}(\omega_k)}{\hat{p}_\mathrm{S}(\omega_k)}}  \,\hat{p}_\mathrm{S}(\omega_k|\bx_i)}{\displaystyle\sum_{k'=1}^K \frac{\hat{p}_\mathrm{T}(\omega_{k'})}{\hat{p}_\mathrm{S}(\omega_{k'})}\,\hat{p}_\mathrm{S}(\omega_{k'}|\bx_i)}
		=
		\frac{\displaystyle{\frac{\theta_k}{\hat{p}_\mathrm{S}(\omega_k)}}  \,\hat{p}_\mathrm{S}(\omega_k|\bx_i)}{\displaystyle\sum_{k'=1}^K \frac{\theta_{k'}}{\hat{p}_\mathrm{S}(\omega_{k'})}\,\hat{p}_\mathrm{S}(\omega_{k'}|\bx_i)}.
\end{equation}
\\
Now by substituting (\ref{LL_full}) in the definition of $Q(\boldsymbol{\theta},\boldsymbol{\theta}^{\mathrm{old}})$ and using (\ref{exp_z_ik}) with $\boldsymbol{\theta}=\boldsymbol{\theta}^{\mathrm{old}}=\boldsymbol{\theta}^{(s)}$ (assuming $s$ iterations have already been performed) we get
\begin{equation}
	\label{QSS_explicit}
	\begin{split}
	Q(\boldsymbol{\theta},\boldsymbol{\theta}^{(s)}) &=
	\sum_{i=1}^m \sum_{k=1}^K \hat{p}^{(s)}_{\mathrm{T}}(\omega_k|\bx_i) \log \hat{p}_\mathrm{T}(\bx_i|\omega_k)
	\\
	&+ 
	\sum_{i=1}^m \sum_{k=1}^K \hat{p}^{(s)}_{\mathrm{T}}(\omega_k|\bx_i) \log \theta_k.
	\end{split}
\end{equation}
The (M) step instructs us to compute $\boldsymbol{\theta} ^{(s+1)}$ as the value of $\boldsymbol{\theta}$ that maximizes (\ref{QSS_explicit}). Remember that the parameters $\theta_k$ represent probabilities, therefore they sum to one and are not independent. We thus construct the Lagrange function:
\begin{equation}
	\widetilde{Q}(\boldsymbol{\theta},\boldsymbol{\theta}^{(s)};\lambda) := 
	Q(\boldsymbol{\theta},\boldsymbol{\theta}^{(s)}) +
	\lambda\left(1-\sum_{k=1}^K \theta_k\right).
\end{equation}
From the vanishing condition for the partial derivatives $\displaystyle \frac{\partial \widetilde{Q}(\boldsymbol{\theta},\boldsymbol{\theta}^{(s)};\lambda)}{\partial \theta_k}$ we readily get:
\begin{equation}
	\label{vanish_part_der}
	\sum_{i=1}^m \hat{p}^{(s)}_{\mathrm{T}}(\omega_k|\bx_i)  = \lambda \theta_k^{(s+1)}.
\end{equation}
Summing (\ref{vanish_part_der}) over $k$ and using the fact that both $\theta_k^{(s+1)}$ and $\hat{p}^{(s)}_{\mathrm{T}}(\omega_k|\bx_i)$ sum to 1, we get $\lambda=m$. Therefore we get the update rule for the next approximation of the target prior values:
\begin{equation}
	\theta_k^{(s+1)} := \hat{p}_\mathrm{T}^{(s+1)}(\omega_k) =  \frac{1}{m}\sum_{i=1}^m \hat{p}^{(s)}_{\mathrm{T}}(\omega_k|\bx_i),
\end{equation}
which is precisely the second equation of the iteration (\ref{prior_fixed_point_iter}). The first equation in the iteration (\ref{prior_fixed_point_iter}) is the first equality in (\ref{p_T_omega_xi_iter}) at iteration $s$. We have thus derived the iteration procedure rigorously as an instance of the EM procedure.

\end{document}